%% file: siamese_reconstruction.tex
\documentclass[10pt,twocolumn,letterpaper]{article}

\usepackage{iccv}
\usepackage{times}
\usepackage{epsfig}
\usepackage{graphicx}
\usepackage{subfigure}
\usepackage{amsmath}
\usepackage{amssymb}
\usepackage{microtype}
\usepackage{multirow}
\usepackage{enumitem}
\usepackage{eso-pic}
\setitemize[0]{leftmargin=10pt}

\newenvironment{tight_itemize}{
\begin{itemize}[leftmargin=10pt]
  \setlength{\topsep}{0pt}
  \setlength{\itemsep}{2pt}
  \setlength{\parskip}{0pt}
  \setlength{\parsep}{0pt}
}{\end{itemize}}

\DeclareMathOperator*{\argmin}{argmin}

\usepackage[pagebackref=true,breaklinks=true,letterpaper=true,colorlinks,bookmarks=false]{hyperref}

\iccvfinalcopy 

\def\iccvPaperID{647} 

\ificcvfinal\fi
\begin{document}

\title{Reconstruction-Based Disentanglement for Pose-invariant Face Recognition}

\author{Xi Peng$^{\dagger}$\thanks{This work was part of the Xi's internship at NEC Laboratories America.}, Xiang Yu$^{\ddag}$, Kihyuk Sohn$^{\ddag}$, Dimitris N. Metaxas$^{\dagger}$ and Manmohan Chandraker$^{\S\ddag}$\\
$^{\dagger}$Rutgers, The State University of New Jersey \\
$^{\S}$University of California, San Diego \\
$^{\ddag}$ NEC Laboratories America \\
{\tt\small \{xipeng.cs, dnm\}@rutgers.edu, \{xiangyu,ksohn,manu\}@nec-labs.com}}

\maketitle
\input{abstract}
\input{introduction}

\input{related_work}
\input{method}

\input{implementation}

\input{experiment}

\input{conclusion}

{\small
\bibliographystyle{ieee}
\bibliography{egbib}
}

\clearpage

\input{supplementary}

\end{document}

%% file: abstract.tex
\begin{abstract}
Deep neural networks (DNNs) trained on large-scale datasets have recently achieved impressive improvements in face recognition. But a persistent challenge remains to develop methods capable of handling large pose variations that are relatively under-represented in training data. This paper presents a method for learning a feature representation that is invariant to pose, without requiring extensive pose coverage in training data. We first propose to generate non-frontal views from a single frontal face, in order to increase the diversity of training data while preserving accurate facial details that are critical for identity discrimination. Our next contribution is to seek a rich embedding that encodes identity features, as well as non-identity ones such as pose and landmark locations. Finally, we propose a new feature reconstruction metric learning to explicitly disentangle identity and pose, by demanding alignment between the feature reconstructions through various combinations of identity and pose features, which is obtained from two images of the same subject. Experiments on both controlled and in-the-wild face datasets, such as MultiPIE, 300WLP and the profile view database CFP, show that our method consistently outperforms the state-of-the-art, especially on images with large head pose variations.~\footnote{Detail results and resource are referred to: \url{https://sites.google.com/site/xipengcshomepage/iccv2017}.} 
\end{abstract}

%% file: introduction.tex
\section{Introduction}
The human visual system is commendable at recognition across variations in pose, for which two theoretical constructs are preferred. The first postulates invariance based on familiarity where separate view-specific visual representations or templates are learned \cite{Edelman_Bulthoff_1992,Poggio_Edelman_1990}. The second suggests that structural descriptions are learned from images that specify relations among viewpoint-invariant primitives \cite{Hummel_Biederman_1992}. Analogously, pose-invariance for face recognition in computer vision also falls into two such categories.

\begin{figure}[t]
\centering
\includegraphics[width=0.96\linewidth]{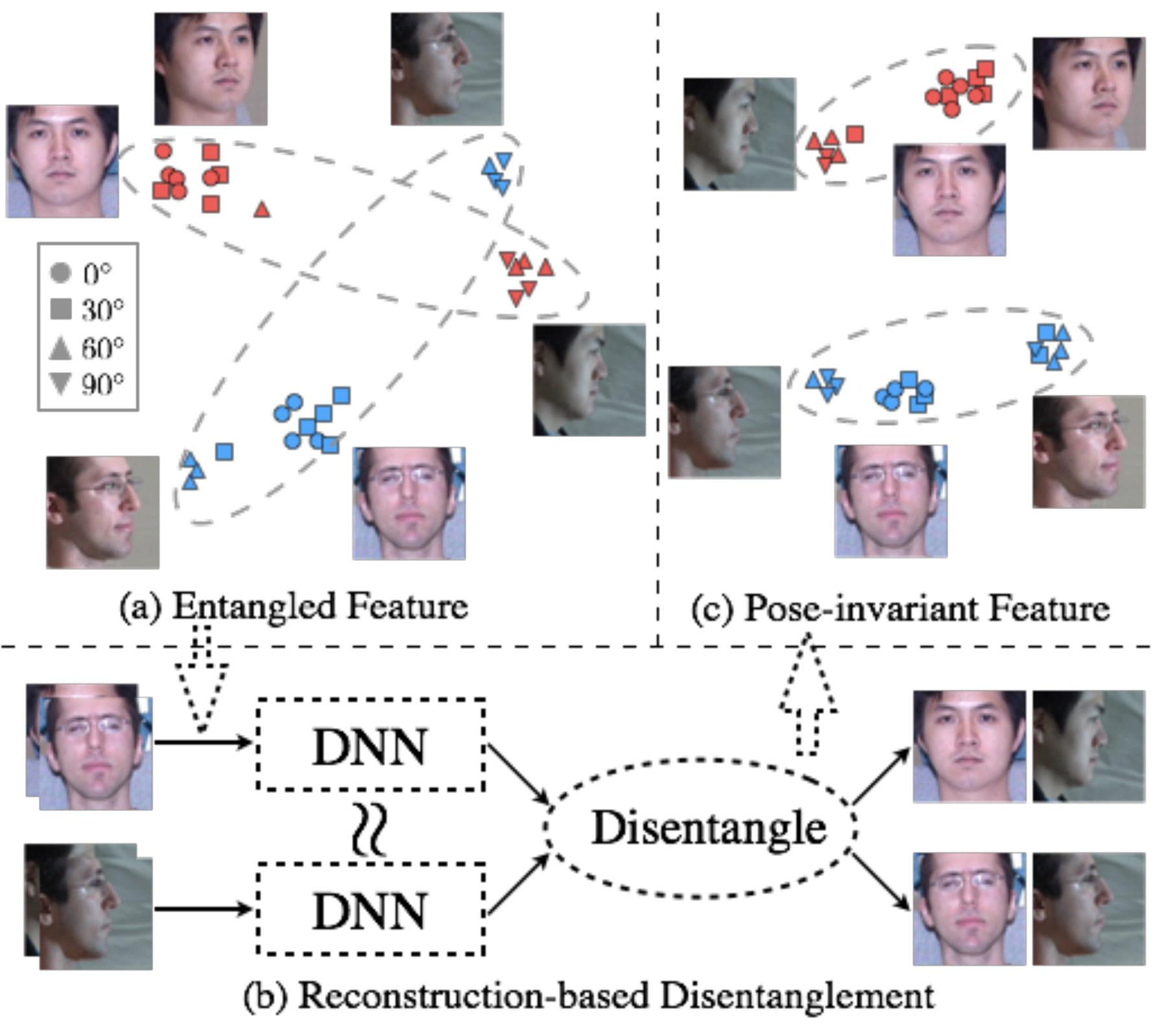}
\caption{(a) Generic data-driven features for face recognition might confound images of the same identity under large poses with other identities, as shown two subjects (in different colors) from MultiPIE are mapped into the learned feature space of VGGFace~\cite{ParkhiBMVC15}. (b) We propose a feature reconstruction metric learning to disentangle identity and pose information in the latent feature space. (c) The disentangled feature space encourages identity features of the same subject to be clustered together despite of the pose variation.}
\label{fig:sample}
\vspace{-3mm}
\end{figure}

The use of powerful deep neural networks (DNNs) \cite{KrizhevskyNIPS12} has led to dramatic improvements in recognition accuracy. However, for objects such as faces where minute discrimination is required among a large number of identities, a straightforward implementation is still ineffective when faced with factors of variation such as pose changes \cite{peng2015circle}. Consider the feature space of the VGGFace \cite{ParkhiBMVC15} evaluated on MultiPIE \cite{multipie} shown in Figure \ref{fig:sample}, where examples from the same identity class that differ in pose are mapped to distant regions of the feature space.
An avenue to address this is by increasing the pose variation in training data. For instance, $4.4$ million face images are used to train DeepFace~\cite{TaigmanCVPR14} and $200$ million labelled faces for FaceNet~\cite{SchroffCVPR15}. Another approach is to learn a mapping from different view-specific feature spaces to a common feature space through methods such as Canonical Correlation Analysis (CCA)~\cite{Hardoon2004}. Yet another direction is to ensemble over view-specific recognition modules that approximate the non-linear pose manifold with locally linear intervals \cite{MasiCVPR16,Kanmeina16}.

There are several drawbacks for the above class of approaches. First, conventional datasets including those sourced from the Internet have long-tailed pose distributions \cite{MasiECCV16}. Thus, it is expensive to collect and label data that provides good coverage for all subjects. Second, there are applications for recognition across pose changes where the dataset does not contain such variations, for instance, recognizing an individual in surveillance videos against a dataset of photographs from identification documents. Third, the learned feature space does not provide insights since factors of variation such as identity and pose might still be entangled. Besides the above limitations, view-specific or multiview methods require extra pose information or images under multiple poses at test time, which may not be available.


In contrast, we propose to learn a novel reconstruction based feature representation that is invariant to pose and does not require extensive pose coverage in training data. A challenge with pose-invariant representations is that discrimination power of the learned feature is harder to preserve, which we overcome with our holistic approach. First, inspired by \cite{ZhuXYCVPR15}, Section \ref{sec:3.1} proposes to enhance the diversity of training data with images under various poses (along with pose labels), at no additional labeling expense, by designing a face generation network. But unlike \cite{ZhuXYCVPR15} which frontalizes non-frontal faces, we {\em generate rich pose variations} from frontal examples, which leads to advantages in better preservation of details and enrichment rather than normalization of within-subject variations. Next, to achieve a rich feature embedding with good discrimination power, Section \ref{sec:3.2} presents a joint learning framework for identification, pose estimation and landmark localization. By jointly optimizing those three tasks, a {\em rich feature embedding} including both identity and non-identity information is learned. But this learned feature is still not guaranteed to be pose-invariant.

To achieve pose invariance, Section \ref{sec:3.3} proposes a feature reconstruction-based structure to explicitly {\em disentangle identity and non-identity} components of the learned feature. The network accepts a reference face image in frontal pose and another image under pose variation and extracts features corresponding to the rich embedding learned above. Then, it minimizes the error between two types of reconstructions in feature space. The first is {\em self-reconstruction}, where the reference sample's identity feature is combined with its non-identity feature and the second is {\em cross-reconstruction}, where the reference sample's non-identity feature is combined with the pose-variant sample's identity feature. This encourages the network to regularize the pose-variant sample's identity feature to be close to that of the reference sample. Thus, non-identity information is distilled away, leaving a disentangled identity representation for recognition at test.

Section \ref{sec:experiments} demonstrates the significant advantages of our approach on both controlled datasets and uncontrolled ones for recognition in-the-wild, especially on $90^{\circ}$ cases. In particular, we achieve strong improvements over state-of-the-art methods on 300-WLP, MultiPIE, and CFP datasets. These improvements become increasingly significant as we consider performance under larger pose variations. We also present ablative studies to demonstrate the utility of each component in our framework, namely pose-variant face generation, rich feature embedding and disentanglement by feature reconstruction.

To summarize, our key contributions are:
\vspace{-0.2cm}
\begin{tight_itemize}
\item To the best of our knowledge, we are the first to propose a novel reconstruction-based feature learning that disentangles factors of variation such as identity and pose.
\item A comprehensively designed framework cascading rich feature embedding with the feature reconstruction, achieving pose-invariance in face recognition.
\item A generation approach to enrich the diversity of training data, without incurring the expense of labeling large datasets spanning pose variations.
\item Strong performance on both controlled and uncontrolled datasets, especially for large pose variations up to $90^{\circ}$.
\end{tight_itemize}

%% file: related_work.tex
\begin{figure*}[t]
\centering
\includegraphics[width=0.9\textwidth]{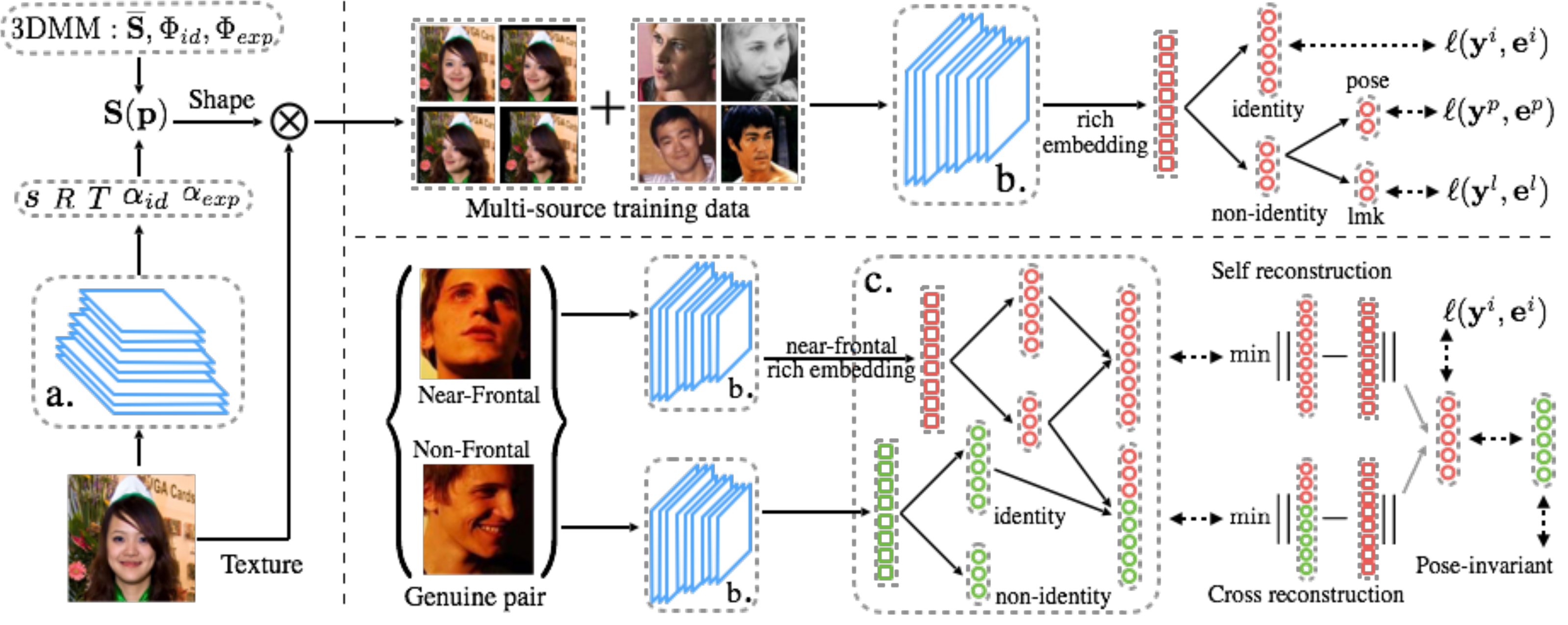}
\caption{An overview of the proposed approach. (a) {\it Pose-variant face generation} utilizes a 3D facial model to synthesize new viewpoints from near-frontal faces. (b) {\it Rich feature embedding} is then achieved by jointly learning the identity and non-identity features using multi-source supervisions. (c) Finally, {\it Disentangling by reconstruction} is applied to distill the identity feature from the non-identity one for robust and pose-invariant representation.} \label{fig:all}
\end{figure*}

\section{Related Work}
While face recognition is an extensively studied area, we provide a brief overview of works most relevant to ours.

\vspace{-3mm}
\paragraph{Face synthesization} Blanz and Vetter pioneered 3D morphable models (3DMM) for high quality face reconstruction \cite{BlanzTPAMI03} and recently, blend shape-based techniques have achieved real-time rates \cite{CaoTVCG14}. For face recognition, such techniques are introduced in DeepFace \cite{TaigmanCVPR14}, where face frontalization is used for enhancing face recognition performance. As an independent application, specific frontalization techniques have also been proposed~\cite{HassnerCVPR15}. Another line of work pertains to 3D face reconstruction from photo collections \cite{roth15,shuliang16,wang2017leveraging} or a single image \cite{MasiECCV16,ZhuXYCVPR15,TranHMM16}, where the latter have been successfully used for face normalization prior to  recognition. While most of the methods apply the framework of aligning 3DMM with the 2D face landmarks~\cite{xiang_cor_2014,xiang_pami_cdm_2015,peng2017toward} and conduct further refinement. In contrast, our use of 3DMM for face synthesis is geared towards enriching the diversity of training data.

\vspace{-3mm}
\paragraph{Deep face recognition} Several frameworks have recently been proposed that use DNNs to achieve impressive performances \cite{ParkhiBMVC15,SchroffCVPR15,SunNIPS2014,SunCVPR14,TaigmanCVPR14,wenyandong2016,YiCoRR14}. DeepFace \cite{TaigmanCVPR14} achieved verification rates comparable to human labeling on large test datasets, with further improvements from works such as DeepID \cite{SunCVPR14}. Collecting face images from the Internet, FaceNet~\cite{SchroffCVPR15} trains on 200 million images from 8 million subjects. The very deep network can only be well stimulated by the huge volume of training data. 
We also use DNNs, but adopt the contrasting approach of learning pose-invariant features, since large-scale datasets with pose variations are expensive to collect, or do not exist in several applications such as surveillance.

\vspace{-3mm}
\paragraph{Pose-invariant face recognition} Early works use Canonical Correlation Analysis (CCA) to analyze the commonality among different pose subspaces \cite{Hardoon2004,nelson2002}. Further works consider generalization across multiple viewpoints \cite{sharma2012} and multiview inter and intra discriminant analysis~\cite{kan2012}. With the introduction of DNNs, prior works aim to transfer information from pose variant inputs to a frontalized appearance~\cite{drgan,xiang_ffgan}, which is then used for face recognition \cite{zhuzhenyao2013}. The frontal appearance reconstruction usually relies on large amount of training data and the pairing across poses is too strict to be practical. Stacked progressive autoencoders (SPAE) \cite{kan2014} map face appearances under larger non-frontal poses to those under smaller ones in a continuous way by setting up hidden layers. The regression based mapping highly depends on training data and may lack generalization ability. Hierarchical-PEP \cite{LiH15cvpr} employs probabilistic elastic part (PEP) model to match facial parts from different yaw angles for unconstrained face recognition scenarios. The 3D face reconstruction method \cite{ZhuXYCVPR15} synthesizes missing appearance due to large view points, which may introduce noise. Rather than compensating the missing information caused by severe pose variations at appearance level, we target learning a pose-invariant representation at feature level which preserves discrimination power through deep training.

\vspace{-3mm}
\paragraph{Disentangle factors of variation}
Contractive discriminative analysis~\cite{rifai2012disentangling} learns disentangled representations in semi-supervised framework by regularizing representations to be orthogonal to each other.
Disentangling Boltzmann machine~\cite{icml2014disentangling} regularizes representations to be specific to each target task via manifold interaction. These methods involve non-trivial training procedure, and the pose variation is limited to half-profile views ($\pm 45^{\circ}$).
Inverse graphics network~\cite{NIPS2015_5851} learns an interpretable representation by learning and decoding graphics codes, each of which encodes different factors of variation, but has been demonstrated only on the database generated from 3D CAD models. 
Multi-View Perceptron~\cite{Zhu14nips} disentangles pose and identity factors by cross-reconstruction of images synthesized from deterministic identity neurons and random hidden neurons. But it does not account for factors such as illumination or expression that are also needed for image-level reconstruction. In contrast, we use carefully designed embeddings as reconstruction targets instead of pixel-level images, which reduces the burden of reconstructing irrelevant factors of variation.

%% file: method.tex
\section{Proposed Method} \label{sec:method}

We propose a novel pose-invariant feature learning method for large pose face recognition. Figure \ref{fig:all} provides an overview of our approach. {\it Pose-variant face generation} utilizes a 3D facial model to augment the training data with faces of novel viewpoints, besides generating ground-truth pose and facial landmark annotations. {\it Rich feature embedding} is then achieved by jointly learning the identity and non-identity features using multi-source supervision. Finally, {\it disentanglement by feature reconstruction} is performed to distill the identity feature from the non-identity one for better discrimination ability and pose-invariance.


\subsection{Pose-variant Face Generation} \label{sec:3.1}

The goal is to generate a series of pose-variant faces from a near-frontal image. This choice of generation approach is deliberate, since it can avoid hallucinating missing textures due to self-occlusion, which is a common problem with former approaches \cite{HassnerCVPR15,DuongCVPR15} that rotate non-frontal faces to a normalized frontal view. More importantly, enriching instead of reducing intra-subject variations provides important training examples in learning pose-invariant features.

We reconstruct the 3D shape from a near-frontal face to generate new face images. Let $\chi$ be the set of frontal face images. A straightforward solution is to learn a nonlinear mapping $f(\cdot; \theta^s): \chi \rightarrow \mathbb{R}^{3N}$ that maps an image $\mathbf{x} \in \chi$ to the $N$ coordinates of a 3D mesh. However, it is non-trivial to do so for a large number of vertices (15k), as required for a high-fidelity reconstruction. 

Instead, we employ the 3D Morphable Model (3DMM) \cite{BlanzTPAMI03} to learn a nonlinear mapping $f(\cdot; \theta^s): \chi \rightarrow \mathbb{R}^{235}$ that embeds $\mathbf{x}$ to a low-dimensional parameter space. The 3DMM parameters $\mathbf{p}$ control the rigid affine transformation and non-rigid deformation from a 3D mean shape $\overline{\mathbf{S}}$ to the instance shape $\mathbf{S}$. Please refer to Figure \ref{fig:all} for an illustration:
\begin{align} \label{eq:3dmm}
	\mathbf{S}(\mathbf{p}) = s R (\overline{\mathbf{S}} + \mathbf{\Phi_{id}} \alpha_{id} + \mathbf{\Phi_{exp}} \alpha_{exp}) + T,
\end{align}
where $\mathbf{p} = \{s, R, T, \alpha_{id}, \alpha_{exp}\}$ including scale $s$, rotation $R$, translation $T$, identity coefficient $\alpha_{id}$ and expression coefficient $\alpha_{exp}$. The eigenbases $\mathbf{\Phi_{id}}$ and $\mathbf{\Phi_{exp}}$ are learned offline using 3D face scans to model the identity \cite{PaysanAVSS09} and expression \cite{CaoTVCG14} subspaces, respectively. 

Once the 3D shape is recovered, we rotate the near-frontal face by evenly manipulating the yaw angle in the range of $[-90^\circ, 90^\circ]$. We follow \cite{ZhuXYCVPR15} to use a z-buffer for collecting texture information and render the background for high-quality recovery. The rendered face is then projected to 2D to generate new face images from novel viewpoints.

\subsection{Rich Feature Embedding} \label{sec:3.2}

\begin{figure}[t]
\centering
\includegraphics[width=0.43\textwidth]{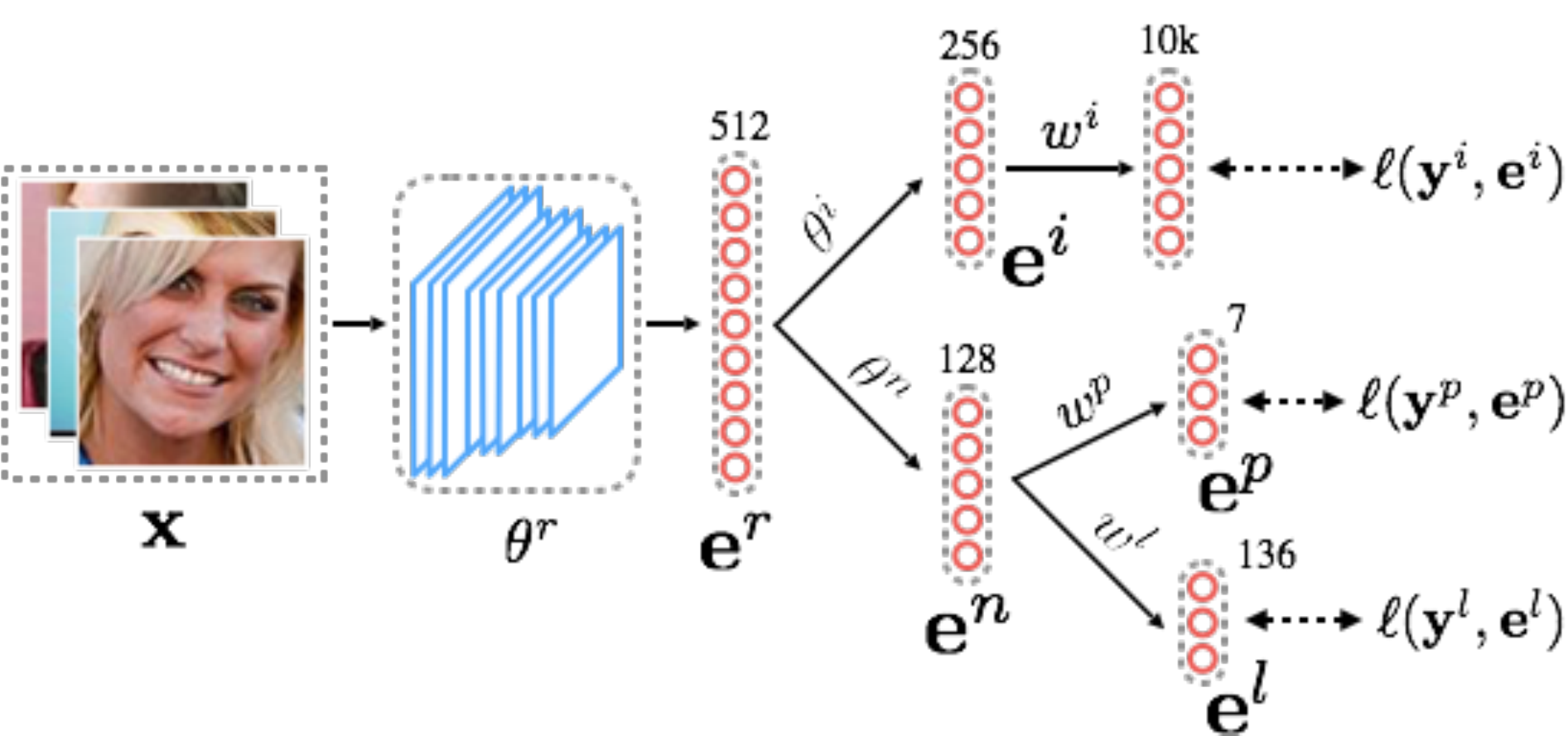}
\caption{Pose-variant faces are used to finetune an off-the-shell recognition network $\theta^r$ to learn the rich feature embedding $\mathbf{e}^r$, which is explicitly branched into the identity feature $\mathbf{e}^i$ and the non-identity feature $\mathbf{e}^n$. Multi-source supervisions, such as identity, pose and landmark, are applied for joint optimization.} \label{fig:mtl}
\end{figure}

Most existing face recognition algorithms \cite{MasiECCV16,MasiCVPR16,SchroffCVPR15,wenyandong2016} learn face representation using only identity supervision. An underlying assumption of their success is that deep networks can ``implicitly" learn to suppress non-identity factors after seeing a large volume of images with identity labels \cite{SchroffCVPR15,TaigmanCVPR14}.

However, this assumption does not always hold when extensive non-identity variations exist. As shown in Figure \ref{fig:sample} (a), the face representation and pose changes still present substantial correlations, even though this representation is learned throught a very deep neural network (VGGFace \cite{ParkhiBMVC15}) with large-scale training data (2.6M).

This indicates that using only identity supervision might not suffice to achieve an invariant representation. Motivated by this observation, we propose to utilize multi-source supervision to learn a rich feature embedding $\mathbf{e}^r$, which can be ``explicitly" branched into an identity feature $\mathbf{e}^{i}$ and a non-identity feature $\mathbf{e}^{n}$, respectively. As we will show in the next section, the two features can collaborate to effectively achieve an invariant representation.

More specifically, as illustrated in Figure \ref{fig:mtl}, $\mathbf{e}^n$ can be further branched as $\mathbf{e}^{p}$ and $\mathbf{e}^{l}$ to represent pose and landmark cues. For our multi-source training data that are not generated, we apply the CASIA-WebFace database~\cite{YiCoRR14} and provide the supervision from an off-the-shelf pose estimator~\cite{xiangeccv16}. Therefore, we have:
\begin{align}
	\mathbf{e}^i & = f(\mathbf{x}; \theta^r, \theta^i), \; \mathbf{e}^n = f(\mathbf{x}; \theta^r, \theta^n),\nonumber \\
	\mathbf{e}^p & = h(\mathbf{e}^{n}; w^p) = f(\mathbf{x}; \theta^r, \theta^{n}, w^p), \nonumber \\
	\mathbf{e}^l & = h(\mathbf{e}^{n}; w^l) = f(\mathbf{x}; \theta^r, \theta^{n}, w^l), \nonumber
\end{align}
where mapping $f(\cdot; \theta/w) : \chi \rightarrow \mathbb{R}^d$ takes ${\mathbf{x}}$ and generates an embedding vector $f({\mathbf{x}})$ and $\theta/w$ denotes the mapping parameters. Here, $\theta^r$ can be any off-the-shelf recognition network. $h(\cdot;\theta)$ is used to bridge two embedding vectors. We jointly learn all embeddings by optimizing:
\begin{align} \label{eq:mtl}
\argmin_{\theta^{r,i,n}, w^{i,p,l}} \sum_{image} & - \lambda^i \big[ \mathbf{y}^i \log softmax( {w^{i}}^T \mathbf{e}^i)) \big] \nonumber \\ 
		 & + \lambda^p \| \mathbf{y}^p - \mathbf{e}^p \|^2_2 + \lambda^l \| \mathbf{y}^l - \mathbf{e}^l \|^2_2,
\end{align}
where $\mathbf{y}^i$, $\mathbf{y}^p$ and $\mathbf{y}^l$ are identity, pose and landmark annotations and $\lambda^i$, $\lambda^p$ and $\lambda^l$ balance the weights between cross-entropy and $l_2$ loss.

By resorting to multi-source supervision, we can learn the rich feature embedding that ``explicitly'' encodes both identity and non-identity cues in $\mathbf{e}^i$ and $\mathbf{e}^n$, respectively. The remaining challenge is to distill $\mathbf{e}^i$ by disentangling from $\mathbf{e}^n$ to achieve identity-only representation.

\subsection{Disentanglement by Feature Reconstruction}
\label{sec:3.3}

The identity and non-identity features above are jointly learned under different supervision. However, there is no guarantee that the identity factor has been fully disentangled from the non-identity one since there is no supervision applied on the decoupling process. This fact motivates us to propose a novel reconstruction-based framework for effective identity and non-identity disentanglement.

Recall that we have generated a series of pose-variant faces for each training subject in Section \ref{sec:3.1}. These images share the same identity but have different viewpoints. We categorize these images into two groups according to their absolute yaw angles: near-frontal faces ($\leq 5^\circ$) and non-frontal faces ($>5^\circ$). The two groups are used to sample image pairs that follow a specially designed configuration: a reference image which is randomly selected from the near-frontal group and a peer image which is randomly picked from the non-frontal group.

\begin{figure}[t]
\centering
\includegraphics[width=0.43\textwidth]{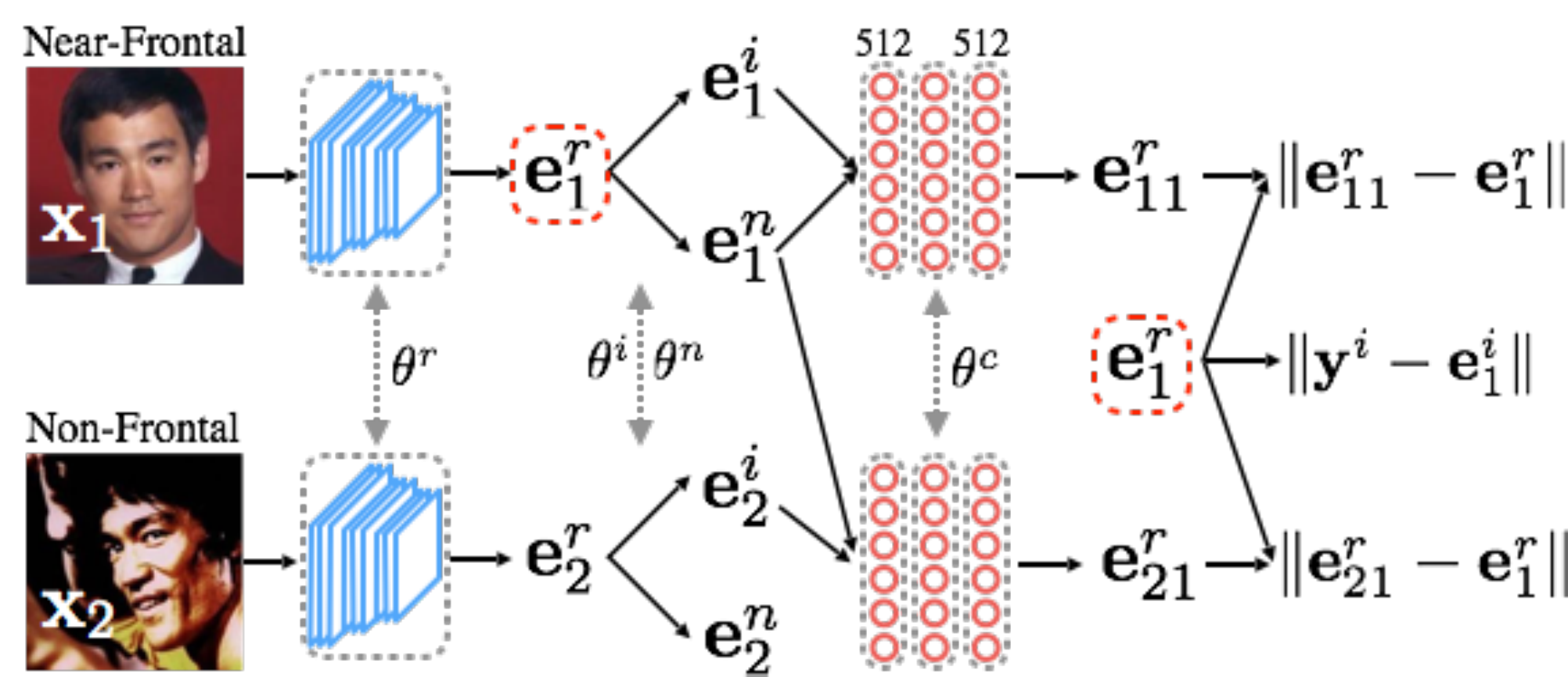}
\caption{ A genuine pair $\{\mathbf{x}_1, \mathbf{x}_2\}$ that share the same identity but different pose is fed into the recognition network $\theta^r$ to obtain the rich embedding $\mathbf{e}^r_1$ and $\mathbf{e}^r_2$. By regularizing the self and cross reconstruction, $\mathbf{e}^r_{11}$ and $\mathbf{e}^r_{21}$, the identity and non-identity features are eventually disentangled to make the non-frontal peer $\mathbf{e}^i_2$ to be similar to its near-frontal reference $\mathbf{e}^i_1$.}
\label{fig:rec}
\end{figure}

The next step is to obtain the identity and non-identity embeddings of two faces that have the same identity but different viewpoints. As shown in Figure \ref{fig:rec}, a pair of images $\{\mathbf{x}_k: k = 1, 2\}$ are fed into the network to output the corresponding identity and non-identity features:   
\begin{align}
	\mathbf{e}^i_k & = f(\mathbf{e}^r_k; \theta^i) = f(\mathbf{x}_k; \theta^r, \theta^i), \nonumber \\
	\mathbf{e}^n_k & = f(\mathbf{e}^r_k; \theta^n) = f(\mathbf{x}_k; \theta^r, \theta^n). \nonumber
\end{align}
Note that $\theta$ is not indexed by $k$ as the network shares weights to process images of the same pair.

Our goal is to eventually push $\mathbf{e}^i_1$ and $\mathbf{e}^i_2$ close to each other to achieve a pose-invariant representation. A simple solution is to directly minimize the $l_2$ distance between the two features in the embedding subspace. However, this constraint only considers the identity branch, which might be entangled with non-identity, but completely ignores the non-identity factor, which provides strong supervision to purify the identity. Our experiments also indicate that a hard constraint would suffer from limited performance in large-pose conditions.

To address this issue, we propose to relax the constraint under a reconstruction-based framework. More specifically, we firstly introduce two reconstruction tasks:
\begin{align}
	\mathbf{e}^r_{11} = g(\mathbf{e}^i_1, \mathbf{e}^n_1; \theta^c), \;
	\mathbf{e}^r_{21} = g(\mathbf{e}^i_2, \mathbf{e}^n_1; \theta^c), \nonumber
\end{align}
where $\mathbf{e}^r_{11}$ denotes the {\it self reconstruction} of the near-frontal rich embedding; while $\mathbf{e}^r_{21}$ denotes the {\it cross reconstruction} of the non-frontal rich embedding.
Here, $g(\cdot, \cdot; \theta^c)$ is the reconstruction mapping with parameter $\theta^c$.

The identity and non-identity features can be rebalanced from the rich feature embedding by minimizing the self and cross reconstruction loss under the cross-entropy constraint:
\begin{align} \label{eq:rec}
\argmin_{\theta^i,\theta^{n},\theta^c} \sum_{pair} & - \gamma^i \big[ \mathbf{y}^i_1 \log softmax( {w^{i}}^T \mathbf{e}^i_1) \big] \nonumber \\ 
		 & + \gamma^s \| \mathbf{e}^r_{11} - \mathbf{e}^r_1 \|^2_2 + \gamma^c \| \mathbf{e}^r_{21} - \mathbf{e}^r_1 \|^2_2,
\end{align}
where $\gamma^i$, $\gamma^s$ and $\gamma^c$ weigh different constraints. Note that compared to \eqref{eq:mtl}, here we only finetune $\{\theta^i, \theta^n\}$ (as well as $\theta^c$) to rebalance the identity and non-identity features while keeping $\theta^r$ fixed, which is an important strategy to maintain the previously learned rich embedding.

In \eqref{eq:rec}, we regularize both self and cross reconstructions to be close to the near-frontal rich embedding $\mathbf{e}^r_1$. Thus, portions of $\mathbf{e}^r_2$ to $\mathbf{e}^i_2$ and $\mathbf{e}^n_2$ are dynamically rebalanced to make the non-frontal peer $\mathbf{e}^i_2$ to be similar to the near-frontal reference $\mathbf{e}^i_1$. In other words, we encourage the network to learn a normalized feature representation across pose variations, thereby disentangling pose information from identity. 

The proposed feature-level reconstruction is significantly different from former methods \cite{SchroffCVPR15,HassnerCVPR15} that attempt to frontalize faces at the image level. It can be directly optimized for pose invariance without suffering from artifacts that are common issues in face frontalization. Besides, our approach is an end-to-end solution that does not rely on extensive preprocessing usually required for image-level face normalization.

Our approach is also distinct from existing methods \cite{MasiCVPR16,MasiECCV16} that synthesize pose-variant faces for data augmentation. Instead of feeding the network with a large number of augmented faces and letting it automatically learn pose-invariant or pose-specific features, we utilize the reconstruction loss to supervise the feature decoupling procedure. Moreover, factors of variation other than pose are also present in training, even though we only use pose as the driver for disentanglement. The cross-entropy loss in \eqref{eq:rec} plays an important role in preserving the discriminative power of identity features across various factors.

%% file: implementation.tex
\section{Implementation Details}


{\bf Pose-variant face generation} A deep network is employed to predict 3DMM parameters of a near-frontal face as shown in Figure \ref{fig:all} (a). The network has a similar architecture as VGG16~\cite{simonyan2014}. We use pre-trained weights learned from ImageNet \cite{KrizhevskyNIPS12} to initialize the network instead of training from scratch. To further improve the performance, we make two important changes: (1) we use stride-2 convolution instead of max pooling to preserve the structure information when halving the feature maps; (2) the dimension of 3DMM parameters is changed to 66-$d$ (30 identity, 29 expression and 7 pose) instead of 235-$d$ used in~\cite{zhuxiangyu2016}.
We evenly sample new viewpoints in every $5^\circ$ from near-frontal faces to left/right profiles to cover the full range of pose variations.



{\bf Rich feature embedding} The network is designed based on CASIA-net~\cite{YiCoRR14} with some improvements. As illustrated in Figure \ref{fig:mtl}, we change the last fully connected layer to 512-$d$ for the rich feature embedding, which is then branched into 256-$d$ neurons for the identity feature and 128-$d$ neurons for the non-identity feature. To utilize multi-source supervision, the non-identity feature is further forked into 7-$d$ neurons for the pose embedding and 136-$d$ neurons for the landmark coordinates. Three different datasets are used to train the network: CASIA-WebFace, 300WLP and MultiPIE. We use Adam \cite{kingma2014adam} stochastic optimizer with an initial learning rate of $0.0003$, which drops by a factor of 0.25 every 5 epochs until convergence. Note that we train the network from scratch on purpose, since a pre-trained recognition model usually has limited ability to re-encode non-identity features.  


{\bf Disentanglement by reconstruction} Once $\{\theta^r,\theta^i,\theta^n\}$ are learned in the rich feature embedding, we freeze $\theta^r$ and finetune $\theta^i$ and $\theta^n$ to rebalance the identity and non-identity features as explained in Figure \ref{fig:rec} and \eqref{eq:rec}. The network takes the concatenation (384-$d$) of $\mathbf{e}^i$ and $\mathbf{e}^n$ and outputs the reconstructed embedding (512-$d$). The mapping is achieved by rolling though two fully connected layers and each of them has 512-$d$ neurons. We have tried different network configurations but get similar performance. The initial learning rate is set to 0.0001 and the hyper-parameters $\gamma^{i,s,c}$ are determined via 5-fold cross-validation. We also find that it is import to do early stopping for effective reconstruction-based regularization. In \eqref{eq:mtl} and \eqref{eq:rec}, we use the cross-entropy loss to preserve the discriminative power of the identity feature. Other identity regularizations, {\it e.g.} triplet loss \cite{SchroffCVPR15}, can be easily applied in a plug-and-play manner.

%% file: experiment.tex
\section{Experiments}\label{sec:experiments}
We evaluate our feature learning method on three main pose-variant databases, MultiPIE~\cite{multipie}, 300WLP~\cite{zhuxiangyu2016} and CFP~\cite{cfp}. We also compare with two top general face recognition frameworks, VGGFace~\cite{ParkhiBMVC15} and N-pair loss face recognition~\cite{sohn2016metric}, and three state-of-the-art pose-invariant face recognition methods, namely, MvDA~\cite{kan2012}, GMA~\cite{sharma2012} and MvDN~\cite{Kanmeina16}. Further, we present an ablation study to emphasize the significance of each module that we carefully designed and a cross-database validation demonstrates the good generalization ability of our method. 

\begin{table}[t]
\begin{center}
\small
\setlength\tabcolsep{2.0pt}
\begin{tabular}{|c|c|c|c|c|c|c|c|}
\hline
Method & $15^{\circ}$ & $30^{\circ}$ & $45^{\circ}$ & $60^{\circ}$ & $75^{\circ}$ & $90^{\circ}$ & Avg\\
\hline\hline
VGGFace~\cite{ParkhiBMVC15} & 0.972 & 0.961 & 0.926 & 0.847 & 0.628 & 0.342 & 0.780\\
\hline
N-pair~\cite{sohn2016metric} & 0.990 & 0.983 & {\bf 0.971} & {\bf 0.944} & 0.811 & 0.468 & 0.861\\
\hline\hline
MvDA~\cite{kan2012}$^{\dagger}$   & \textbf{1.000} & 0.979 & 0.909 & 0.855 & 0.718 & 0.564 & 0.837\\
\hline
GMA~\cite{sharma2012}$^{\dagger}$    & \textbf{1.000} & \textbf{1.000} & 0.904 & 0.852 & 0.725 & 0.550 & 0.838\\
\hline
MvDN~\cite{Kanmeina16}$^{\dagger}$   & \textbf{1.000}  & 0.991 & 0.921 & 0.897 & 0.810 & 0.706 & 0.887\\
\hline\hline

Ours (P1) & 0.972 & 0.966 & 0.956 & 0.927 & \textbf{0.857} & \textbf{0.749} & \textbf{0.905}\\
\hline
Ours (P2) & 1.000 & 1.000 & 0.995 & 0.982 & 0.931 & 0.817 & 0.954\\

\hline
\end{tabular}
\end{center}
\caption{Rank-1 recognition accuracy on MultiPIE at different yaw angles. The numbers in the entry with $^{\dagger}$ are obtained from~\cite{Kanmeina16}. We evaluate our method using gallery set composed of 2 frontal face images per subject (P1) as well as entire frontal face images (P2).}
\label{tb:mpie}
\vspace{-4mm}
\end{table}

\begin{table}[t]
\begin{center}
\small
\setlength\tabcolsep{2.0pt}
\begin{tabular}{|c|c|c|c|c|c|c|c|}
\hline
Method & $15^{\circ}$ & $30^{\circ}$ & $45^{\circ}$ & $60^{\circ}$ & $75^{\circ}$ & $90^{\circ}$ & Avg\\
\hline \hline
VGGFace~\cite{ParkhiBMVC15} & 0.994 & 0.998 & \textbf{0.996} & 0.956 & 0.804 & 0.486 & 0.838\\
\hline
N-Pair~\cite{sohn2016metric} &  \textbf{1.000} & 0.996 & 0.993 & 0.962 & 0.845 & 0.542 & 0.859\\
\hline \hline
Ours           &  \textbf{1.000} & \textbf{0.999} & 0.995 & \textbf{0.994} & \textbf{0.978} & \textbf{0.940} & \textbf{0.980}\\
\hline
\end{tabular}
\end{center}
\caption{Recognition performance on 300WLP, the proposed method with two general state-of-the-art face recognition frameworks, i.e. VGG Face Recognition Network (VGGFace) and N-pair loss face recognition (N-pair).}
\label{tb:300wlp}
\vspace{-4mm}
\end{table}

\begin{table*}[t]
\begin{center}
\small
\setlength\tabcolsep{3pt}
\begin{tabular}{|c|c|c|c|c|c|c|c||c|c|c|c|c|c|c|}
\hline
\multirow{2}{*}{Method} & \multicolumn{7}{c||}{MultiPIE} & \multicolumn{7}{c|}{300WLP}\\
\cline{2-15}
& $15^{\circ}$ & $30^{\circ}$ & $45^{\circ}$ & $60^{\circ}$ & $75^{\circ}$ & $90^{\circ}$ & Avg & $15^{\circ}$ & $30^{\circ}$ & $45^{\circ}$ & $60^{\circ}$ & $75^{\circ}$ & $90^{\circ}$ & Avg\\
\hline\hline
SS  & 0.908 & 0.899 & 0.864 & 0.778 & 0.487 & 0.207 & 0.690 & 0.945 & 0.934 & 0.884 & 0.753 & 0.567 & 0.330 & 0.679\\
\hline
SS-FT & 0.941 & 0.936 & 0.919 & 0.883 & 0.799 & 0.681 & 0.860 & \textbf{1.000} & 0.999 & 0.992 & 0.973 & 0.934 & 0.839 & 0.944\\
\hline
MSMT   & 0.965 & 0.955 & 0.945 & 0.914 & 0.827 & 0.689 & 0.882 & \textbf{1.000} & 0.993 & 0.993 & 0.986 & 0.968 & 0.922 & 0.971\\
\hline
MSMT+L2  & 0.972 & 0.965 & 0.954 & 0.923 & 0.849 & 0.739 & 0.900 & \textbf{1.000} & 0.997 & 0.996 & 0.991 & 0.973 & 0.933 & 0.977\\
\hline
MSMT+SR        & 0.972 & 0.966 & 0.956 & 0.927 & 0.857 & \textbf{0.749} & 0.905 & \textbf{1.000} & \textbf{0.999} & 0.995 & 0.994 & 0.978 & 0.940 & 0.980\\
\hline
MSMT$^{\dagger}$ & 0.993 & 0.989 & 0.982 & 0.959 & 0.903 & 0.734 & 0.927 & \textbf{1.000} & 0.998 & 0.997 & 0.994 & 0.981 & 0.922 & 0.977\\
\hline
MSMT$^{\dagger}$+SR & \textbf{0.994} & \textbf{0.990} & \textbf{0.982} & \textbf{0.960} & \textbf{0.906} & 0.745 & \textbf{0.929} & \textbf{1.000}  & 0.998 & \textbf{0.999} & \textbf{0.997} & \textbf{0.988}  & \textbf{0.953} & \textbf{0.986} \\
\hline
\end{tabular}
\end{center}
\caption{Recognition performance of several baseline models, i.e., single source trained model on CASIA database (SS), single source model fine-tuned on the target database (SS-FT), multi-source multi-task models (MSMT), MSMT with direct identity feature $\ell_{2}$ distance regularization (MSMT+L2), the proposed MSMT with Siamese reconstruction regularization models (MSMT+SR), MSMT with N-pair loss instead of cross entropy loss (MSMT$^\dagger$) and MSMT$^\dagger$ with SR, evaluated on MultiPIE (P1) and 300WLP.}
\label{tb:module_compare}
\vspace{-2mm}
\end{table*}
\subsection{Evaluation on MultiPIE}
MultiPIE~\cite{multipie} is composed of 754,200 images of 337 subjects with different factors of variation such as pose, illumination, and expression.
There are 15 different head poses set up, where we only use images of 13 head poses with yaw angle changes from $-90^{\circ}$ to $90^{\circ}$, with $15^{\circ}$ difference every consecutive pose bin in this experiment.

We split the data into train and test by subjects, of which the first 229 subjects are used for training and the remaining 108 are used for testing. This is similar to the experimental setting in~\cite{Kanmeina16}, but we use entire data including both illumination and expression variations for training while excluding only those images taken with top-down views.
Rank-1 recognition accuracy of non-frontal face images is reported. We take $\pm 15^{\circ}$ to $\pm 90^{\circ}$ as query and the frontal faces ($0^{\circ}$) as gallery, while restricting illumination condition to be neutral.

To be consistent with the experimental setting of~\cite{Kanmeina16}, we form a gallery set by randomly selecting 2 frontal face images per subject, of which there are a total of 216 images. We evaluate the recognition accuracy for all query examples, of which there are 619 images per pose.
The procedure is done with 10 random selections of gallery sets and mean accuracy is reported.

Evaluation is shown in Table~\ref{tb:mpie}. 
The recognition accuracy at every $15^{\circ}$ interval of yaw angle is reported while averaging its symmetric counterpart with respect to the 0-yaw axis.
For the two general face recognition algorithms, VGGFace~\cite{ParkhiBMVC15} and N-pair loss~\cite{sohn2016metric}, we clearly observe more than 30\% accuracy drop when the head pose approaches $90^{\circ}$ from $75^{\circ}$. Our method significantly reduces the drop by more than 20\%. The general methods are trained with very large databases leveraging across different poses, but our method has the additional benefit of explicitly aiming for a pose invariant feature representation.


The pose-invariant methods, GMA, MvDA, and MvDN demonstrate good performance within $30^{\circ}$ yaw angles, but again the performance starts to degrade significantly when yaw angle is larger than $30^{\circ}$. When comparing the accuracy on extreme poses from $45^{\circ}$ to $90^{\circ}$, our method achieves accuracy $3\sim4\%$ better than the best reported.
Besides the improved performance, our method has an advantage over MvDN, since it does not require pose information at test time. On the other hand, MvDN is composed of multiple sub-networks, each of which is specific to a certain pose variation and therefore requires additional information on head pose for recognition.


\begin{table}[t]
\begin{center}
\small
\setlength\tabcolsep{3pt}
\begin{tabular}{|c|c|c|}
\hline
Method & Frontal-Frontal & Frontal-Profile\\
\hline
\hline
Sengupta et al.~\cite{cfp} & 96.40 & 84.91 \\
\hline
Sankarana et al.~\cite{sanka2016} & 96.93 & 89.17 \\
\hline
Chen et al.~\cite{chenicip2016} & \textbf{98.67} & 91.97 \\
\hline
DR-GAN~\cite{drgan} & 97.84 & 93.41 \\
\hline
\hline
Human & 96.24 & 94.57 \\
\hline
\hline
Ours & \textbf{98.67} & \textbf{93.76}\\
\hline
\end{tabular}
\end{center}
\caption{Verification accuracy comparison on CFP dataset.}
\label{tb:cfp}
\vspace{-4mm}
\end{table}
\subsection{Evaluation on 300WLP}
We further evaluate on a face-in-the-wild database, 300 Wild Large Pose~\cite{zhuxiangyu2016} (300WLP). It is generated from 300W~\cite{300w2013} face database by 3DDFA~\cite{zhuxiangyu2016}, in which it establishes a 3D morphable model and reconstruct the face appearance with varying head poses. It consists of overall 122,430 images from 3,837 subjects. Compared to MultiPIE, the overall volume is smaller, but the number of subjects is significantly larger. For each subject, images are with uniformly distributed continuously varying head poses in contrast to MultiPIE's strictly controlled $15^{\circ}$ head pose intervals. The lighting conditions as well as the background are almost identical. Thus, it is an ideal dataset to evaluate algorithms for pose variation.

\begin{table*}[t]
\begin{center}
\small
\setlength\tabcolsep{3pt}
\begin{tabular}{|c|c|c|c|c|c|c|c|c||c|c|c|c|c|c|c|}
\hline
\multicolumn{2}{|c|}{\multirow{2}{*}{Method}} & \multicolumn{7}{c||}{MultiPIE} & \multicolumn{7}{c|}{300WLP}\\
\cline{3-16}
\multicolumn{2}{|c|}{}& $15^{\circ}$ & $30^{\circ}$ & $45^{\circ}$ & $60^{\circ}$ & $75^{\circ}$ & $90^{\circ}$ & Avg & $15^{\circ}$ & $30^{\circ}$ & $45^{\circ}$ & $60^{\circ}$ & $75^{\circ}$ & $90^{\circ}$ & Avg\\
\hline \hline
\multirow{2}{*}{MultiPIE}& MSMT   & 0.965 & 0.955 & 0.945 & 0.914 & 0.827 & 0.689 & 0.882 & 1.000 & 0.996 & 0.988 & 0.953 & 0.889 & 0.720 & 0.904\\
\cline{2-16}
& Ours & {0.972} & {0.966} & {0.956} & {0.927} & {0.857} & {0.749} & {0.905} & 0.994 & 0.995 & 0.992 & 0.958 & 0.901 & 0.733 & 0.910\\
\hline
\multirow{2}{*}{300WLP} & MSMT & 0.941 & 0.927 & 0.898 & 0.837 & 0.695 & 0.432 & 0.788 & {1.000} & 0.993 & 0.993 & 0.986 & 0.968 & 0.922 & 0.971\\
\cline{2-16}
 & Ours & 0.945 & 0.933 & 0.910 & 0.862 & 0.736 & 0.459 & 0.808 & {1.000} & {0.999} & {0.995} & {0.994} & {0.978} & {0.940} & {0.980}\\
\hline
\end{tabular}
\end{center}
\caption{Cross database evaluation on MultiPIE and 300WLP. The top two rows show the model of MSMT and our method trained on CASIA and MultiPIE, while tested on both MultiPIE and 300WLP. The bottom two rows show the model of MSMT and our method trained on CASIA and 300WLP, while tested on both MultiPIE and 300WLP.}
\label{tb:cross_model}
\vspace{-4mm}
\end{table*}

We randomly split 500 subjects of 8014 images as testing data and the rest 3337 subjects of 106,402 images as the training data. Among the testing data, two $0^{\circ}$ head pose images per subject form the gallery and the rest 7014 images serves as the probe. Table~\ref{tb:300wlp} shows the comparison with two state-of-the-art general face recognition methods, i.e. VGGFace~\cite{ParkhiBMVC15} and N-pair loss face recognition~\cite{sohn2016metric}. To the best of our knowledge, we are the first to apply our pose-invariant face recognition framework on this dataset. Thus, we only compare our method with the two general face recognition frameworks. 

Since head poses in 300WLP continuously vary, we group the test samples into 6 pose intervals, $(0,15^{\circ})$, $(15^{\circ}, 30^{\circ})$, $(30^{\circ}, 45^{\circ})$, $(45^{\circ}, 60^{\circ})$, $(60^{\circ}, 75^{\circ})$ and $(75^{\circ}, 90^{\circ})$. For short annotation, we mark each interval with the end point, e.g., $30^{\circ}$ denotes the pose interval $(15^{\circ}, 30^{\circ})$. From Table~\ref{tb:300wlp}, our method achieves consistently better accuracy especially when pose angle approaches $90^{\circ}$, which is clearly contributed by our feature reconstruction based disentanglement.



\subsection{Evaluation on CFP}
The Celebrities in Frontal-Profile (CFP) database~\cite{cfp} focuses on extreme head pose face verification. It consists of 500 subjects, with 10 frontal images and 4 profile images for each, in a wild setting. The evaluation is conducted by averaging the performance of 10 randomly selected splits with 350 identical and 350 non-identical pairs. Our MSMT+SR finetuned on MultiPIE with N-pair loss is the model evaluated in this experiment. The reported human performance is 94.57\% accuracy on the frontal-profile protocol and 96.24\% on the frontal-frontal protocol, which shows the challenge of recognizing profile views.

Results in Table \ref{tb:cfp} suggest that our method achieves consistently better performance compared to state-of-the-art. We reach the same Frontal-Frontal accuracy as Chen et al.~\cite{chenicip2016} while being significantly better on Frontal-Profile by 1.8\%. We are slightly better than DR-GAN~\cite{drgan} on extreme pose evaluation and 0.8\% better on frontal cases. DR-GAN is a recent generative method that seeks the identity preservation at the image level, which is not a direct optimization on the features. Our feature reconstruction method preserves identity even when presented with profile view faces. In particular, as opposed to prior methods, ours is the only one that obtains very high accuracy on both the evaluation protocols.  

\vspace{-0.05in}
\subsection{Control Experiments}
\vspace{-0.05in}
We extensively evaluate recognition performance on various baselines to study the effectiveness of each module in our proposed framework.
Specifically, we evaluate and compare the following models: 
\begin{itemize}[itemsep=5pt,nolistsep]
\item{SS: trained on a single source (e.g., CASIA-WebFace) using softmax loss only.}
\item{SS-FT: SS fine-tuned on a target dataset (e.g., MultiPIE or 300WLP) using softmax loss only.}
\item{MSMT: trained on multiple data sources (e.g., CASIA + MultiPIE or 300WLP) using softmax loss for identity and $L_2$ loss for pose.}
\item{MSMT+L2: fine-tuned on MSMT models using softmax loss and Euclidean loss on pairs.}
\item{MSMT+SR: fine-tuned on MSMT models using softmax loss and Siamese reconstruction loss.}
\item{MSMT$^{\dagger}$: trained on the same multiple data sources as MSMT, using N-pair~\cite{sohn2016metric} metric loss for identity and $L_2$ loss for pose.}
\item{MSMT$^{\dagger}$+SR: finetuned on MSMT$^{\dagger}$ models with N-pair loss and reconstruction loss.}
\end{itemize}
The SS model serves as the weakest baseline.
We observe that simultaneously training the network on multiple sources of CASIA and MultiPIE (or 300WLP) using multi-task objective (i.e., identification loss, pose or landmark estimation loss) is more effective than single-source training followed by fine-tuning. 
We believe that our MSMT learning can be viewed as a form of curriculum learning~\cite{BengioLCW09} since multiple objectives introduced by multi-source and multi-task learning are at different levels of difficulty (e.g., pose and landmark estimation or identification on MultiPIE and 300WLP are relatively easier than identification on CASIA-WebFace) and easier objectives allow to train faster and converge to better solution.

As an alternative to reconstruction regularization, one may consider reducing the distance between the identity-related features of the same subject under different pose directly (MSMT+L2). 
Learning to reduce the distance improves the performance over the MSMT model, but is not as effective as our proposed reconstruction regularization method, especially on face images with large pose variations.


Further, we observe that employing the N-pair loss~\cite{sohn2016metric} within our framework also boosts performance, which is shown by the improvements from MSMT to MSMT$^{\dagger}$ and MSMT+SR to MSMT$^{\dagger}$+SR. We note that the MSMT$^{\dagger}$ baseline is not explored in prior works on pose-invariant face recognition. It provides a different way to achieve similar goals as the proposed reconstruction method. Indeed, a collateral observation through the relative performances of MSMT and MSMT$^{\dagger}$ is that the softmax loss is not good at disentangling pose from identity, while metric learning excels at it. Indeed, our feature reconstruction metric might be seen as achieving a similar goal, thus, improvements over MSMT$^{\dagger}$ are marginal, while those over MSMT are large.

\vspace{-0.05in}
\subsection{Cross Database Evaluation}
\vspace{-0.05in}
We evaluate our models, which are trained on CASIA with MultiPIE or 300WLP, on the cross test set 300WLP or MultiPIE, respectively. Results are shown in Table~\ref{tb:cross_model} to validate the generalization ability. There are obvious accuracy drops on both databases, for instance, a $7\%$ drop on 300WLP and $10\%$ drop on MultiPIE. However, such performance drops are expected since there exists a large gap in the distribution between MultiPIE and 300WLP.

Interestingly, we observe significant improvements when compared to VGGFace. These are fair comparisons since neither networks is trained on the training set of the target dataset. 
When evaluated on MultiPIE, our MSMT model trained on 300WLP and CASIA database improves $0.8\%$ over VGGFace and the model with reconstruction regularization demonstrates stronger performance, showing $2.8\%$ improvement over VGGFace.
Similarly, we observe $6.6\%$ and $7.2\%$ improvements for MultiPIE and CASIA trained MSMT models and our proposed MSMT+SR, respectively, over VGGFace when evaluated on the 300WLP test set.
This partially confirms that our performance is not an artifact of overfitting to a specific dataset, but is generalizable across different datasets of unseen images.

%% file: conclusion.tex
\vspace{-0.1in}
\section{Conclusion}
In the paper, we propose a new reconstruction loss to regularize identity feature learning for face recognition. 
We also introduce a data synthesization strategy to enrich the diversity of pose, requiring no additional training data. Rich embedding has already shown promising effects revealed by our control experiments, which is interpreted as curriculum learning.
The self and cross reconstruction regularization achieves successful disentanglement of identity and pose, to show significant improvements on both MultiPIE, 300WLP and CFP with $2\%$ to $12\%$ gaps. Cross-database evaluation further verifies that our model generalizes well across databases. Future work will focus on closing the systematic gap among databases and further improve the generalization ability.

%% file: supplementary.tex
\twocolumn[{%
 \centering
 \Large {\bf Reconstruction-Based Disentanglement for Pose-invariant Face Recognition\\Supplementary Material} \\[1em]
 \large Xi Peng, 
 		Xiang Yu, 
 		Kihyuk Sohn, 
 		Dimitris N. Metaxas 
 	and Manmohan Chandraker \\[1em]
}]

\setcounter{section}{0}

\section{Summary of The Supplementary}
This supplementary file includes two parts: {\bf (a)} Additional implementation details are presented to improve the reproducibility; {\bf (b)} More experimental results are presented to validate our approach in different aspects, which are not shown in the main submission due to the space limitation.

\section{Additional Implementation Details}

{\bf Pose-variant face generation} We designed a network to predict 3DMM parameters from a single face image. The design is mainly based on VGG16~\cite{ParkhiBMVC15}. We use the same number of convolutional layers as VGG16 but replacing all max pooling layers with stride-2 convolutional operations. The fully connected (fc) layers are also different: we first use two fc layers, each of which has 1024 neurons, to connect with the convolutional modules; then, a fc layer of 30 neurons is used for identity parameters, a fc layer of 29 neurons is used for expression parameters, and a fc layer of 7 neurons is used for pose parameters. Different from \cite{ZhuXYCVPR15} uses 199 parameters to represent the identity coefficients, we truncate the number of identity eigenvectors to 30 which preserves $90\%$ of variations. This truncation leads to fast convergence and less overfitting. For texture, we only generate non-frontal faces from frontal ones, which significantly mitigate the hallucinating texture issue caused by self occlusion and guarantee high-fidelity reconstruction. We apply the Z-Buffer algorithm used in ~\cite{ZhuXYCVPR15} to prevent ambiguous pixel intensities due to same image plane position but different depths. 

{\bf Rich feature embedding} The design of the rich embedding network is mainly based on the architecture of CASIA-net~\cite{YiCoRR14} since it is wildly used in former approach and achieves strong performance in face recognition. During training, CASIA+MultiPIE or CASIA+300WLP are used. As shown in Figure 3 of the main submission, after the convolutional layers of CASIA-net, we use a 512-$d$ FC for the rich feature embedding, which is further branched into a 256-$d$ identity feature and a 128-$d$ non-identity feature. The 128-$d$ non-identity feature is further connected with a 136-d landmark prediction and a 7-$d$ pose prediction. Notice that in the face generation network, the number of pose parameters is 7 instead of 3 because we need to uniquely depict the projection matrix from the 3D model and the 2D face shape in image domain, which includes scale, pitch, yaw, roll, x translation, y translation, and z translations.

{\bf Disentanglement by feature reconstruction} Once the rich embedding network is trained, we feed genius pair that share the same identity but different viewpoints into the network to obtain the corresponding rich embedding, identity and non-identity features. To disentangle the identity and pose factors, we concatenate the identity and non-identity features and roll though two 512-$d$ fully connected layers to output a reconstructed rich embedding depicted by 512 neurons. Both self and cross reconstruction loss are designed to eventually push the two identity features close to each other. At the same time, a cross-entropy loss is applied on the near-frontal identity feature to maintain the discriminative power of the learned representation. The disentanglement of the identity and pose is finally achieved by the proposed feature reconstruction based metric learning.

\section{Additional Experimental Results}
In addition to the main submission, we present more experimental results in this section to further validate our approach in different aspects.


\begin{table*}[th]
\begin{center}
\small
\setlength\tabcolsep{4.5pt}
\begin{tabular}{|c|c|c|c|c|c|c|c|}
\hline
\multirow{2}{*}{Method} & \multicolumn{7}{c|}{MultiPIE}\\
\cline{2-8}
& $15^{\circ}$ & $30^{\circ}$ & $45^{\circ}$ & $60^{\circ}$ & $75^{\circ}$ & $90^{\circ}$ & Avg \\
\hline\hline
SS  & 0.908(0.0088) & 0.899(0.0088) & 0.864(0.0072) & 0.778(0.0084) & 0.487(0.0119) & 0.207(0.0156) & 0.690(0.2600) \\
\hline
SS-FT & 0.941(0.0067) & 0.936(0.0090) & 0.919(0.0105) & 0.883(0.0113) & 0.799(0.0108) & 0.681(0.0130) & 0.860(0.0940) \\
\hline
MSMT   & 0.965(0.0053) & 0.955(0.0054) & 0.945(0.0062) & 0.914(0.0059) & 0.827(0.0110) & 0.689(0.0143) & 0.882(0.0982) \\
\hline
MSMT+L2  & 0.972(0.0058) & 0.965(0.0056) & 0.954(0.0075) & 0.923(0.0048) & 0.849(0.0067) & 0.739(0.0095) & 0.900(0.0834) \\
\hline
MSMT+SR (ours) & \textbf{0.972(0.0060)} & \textbf{0.966(0.0069)} & \textbf{0.955(0.0068)} & \textbf{0.927(0.0068)} & \textbf{0.857(0.0066)} & \textbf{0.749(0.0105)} & \textbf{0.905(0.0797)} \\
\hline
\end{tabular}
\end{center}
\caption{Rank-1 recognition accuracy comparisons on MultiPIE~\cite{multipie} under P1 testing protocol.}
\label{tb:module_mpie}
\end{table*}

\begin{table*}[th]
\begin{center}
\small
\setlength\tabcolsep{3pt}
\begin{tabular}{|c|c|c|c|c|c|c|c|c|}
\hline
\multicolumn{2}{|c|}{\multirow{2}{*}{Method}} & \multicolumn{7}{c|}{MultiPIE}\\
\cline{3-9}
\multicolumn{2}{|c|}{}& $15^{\circ}$ & $30^{\circ}$ & $45^{\circ}$ & $60^{\circ}$ & $75^{\circ}$ & $90^{\circ}$ & Avg\\
\hline \hline
\multirow{2}{*}{300WLP}& MSMT (P1)   & 0.941(0.0051) & 0.927(0.0059) & 0.898(0.0073) & 0.837(0.0106) & 0.695(0.0135) & 0.432(0.0110) & 0.788(0.1794)\\
\cline{2-9}
& Ours (P1)& \textbf{0.945(0.0067)} & \textbf{0.933(0.0068)} & \textbf{0.910(0.0073)} & \textbf{0.862(0.0082)} & \textbf{0.736(0.0096)} & \textbf{0.459(0.01359)} & \textbf{0.808(0.1709)} \\
\hline

\hline
\multirow{2}{*}{300WLP}& MSMT (P2)   & 1.00 & 1.00 & 0.992 & 0.943 & 0.797 & 0.488 & 0.870\\
\cline{2-9}
& Ours (P2)& {1.00} & {1.00} & \textbf{0.993} & \textbf{0.964} & \textbf{0.838} & \textbf{0.511} & \textbf{0.884} \\
\hline
\end{tabular}
\end{center}
\caption{Cross database evaluation under either P1 or P2 protocols. Training: CASIA~\cite{YiCoRR14} and 300WLP~\cite{ZhuXYCVPR15}. Testing: MultiPIE~\cite{multipie}.}
\label{tb:cross_model}
\end{table*}

\subsection{P1 and P2 protocol on MultiPIE}
In the main submission, due to space considerations, we only report the mean accuracy over 10 random training and testing splits, on MultiPIE and 300WLP separately. In Table~\ref{tb:module_mpie}, we report the standard deviation of our method as a more complete comparison. From the results, the standard deviation of our method is also very small, which suggests that the performance is consistent across all the trials. We also compare the cross database evaluation on both mean accuracy and standard deviation in Table~\ref{tb:cross_model}. We show the models trained on 300WLP and tested on MultiPIE with both P1 and P2 protocol. Please note that with P2 protocol, our method still achieves better performance on MultiPIE than MvDN~\cite{Kanmeina16} with 0.7\% gap. Further, across different testing protocols, the proposed method consistently outperforms the baseline method MSMT, which clearly shows the effectiveness of our proposed Siamese reconstruction based regularization for pose-invariant feature representation.


\subsection{Control Experiments with P2 on MultiPIE}
The P2 testing protocol utilizes all the $0^{\circ}$ images as the gallery. The performance is expected to be better than that reported on P1 protocol in the main submission since more images are used for reference. There is no standard deviation in this experiment as the gallery is fixed by using all the frontal images. The results are shown in Table~\ref{tb:module_compare}, which confirms the conclusion that the proposed feature reconstruction based regularization is effective in obtaining pose-invariant and highly discriminative feature representations for face recognition.

\begin{table}[t]
\begin{center}
\small
\setlength\tabcolsep{2.5pt}
\begin{tabular}{|c|c|c|c|c|c|c|c|}
\hline
\multirow{2}{*}{Method} & \multicolumn{7}{c|}{MultiPIE}\\
\cline{2-8}
& $15^{\circ}$ & $30^{\circ}$ & $45^{\circ}$ & $60^{\circ}$ & $75^{\circ}$ & $90^{\circ}$ & Avg \\
\hline\hline
SS  & 1.00 & 0.998 & 0.985 & 0.892 & 0.563 & 0.250 & 0.781 \\
\hline
SS-FT & 0.999 & 0.993 & 0.981 & 0.951 & 0.874 & 0.753 & 0.925 \\
\hline
MSMT   & 1.00 & 1.00 & 0.993 & 0.982 & 0.908 & 0.753 & 0.939 \\
\hline
MSMT+L2  & 1.00 & 999 & 0.990 & 0.978 & 0.911 & 0.800 & 0.946 \\
\hline
MSMT+SR (ours) & \textbf{1.00} & 0.999 & \textbf{0.995} & \textbf{0.982} & \textbf{0.931} & \textbf{0.817} & \textbf{0.954} \\
\hline
\end{tabular}
\end{center}
\caption{Recognition accuracy of different baseline models.}
\label{tb:module_compare}
\end{table}


\subsection{Recognition Accuracy on LFW}
We also carried out additional experiments on LFW \cite{lfwdatabase}. As we know, LFW contains mostly near-frontal faces. To better reveal the contribution of our method designed to regularize pose variations, we compare the performance with respect to statistics of pose range (correct pairs num. / total pairs num. in the range). Table~\ref{tb:lfw} shows the results. Our approach outperforms VGG-Face especially in non-frontal settings (>30), which demonstrates the effectiveness of the proposed method in handling pose variations.


\subsection{Feature Embedding of MultiPIE}
\begin{figure*}[t]
\centering
\subfigure[VGGFace Feature Space]{\includegraphics[width=0.46\linewidth]{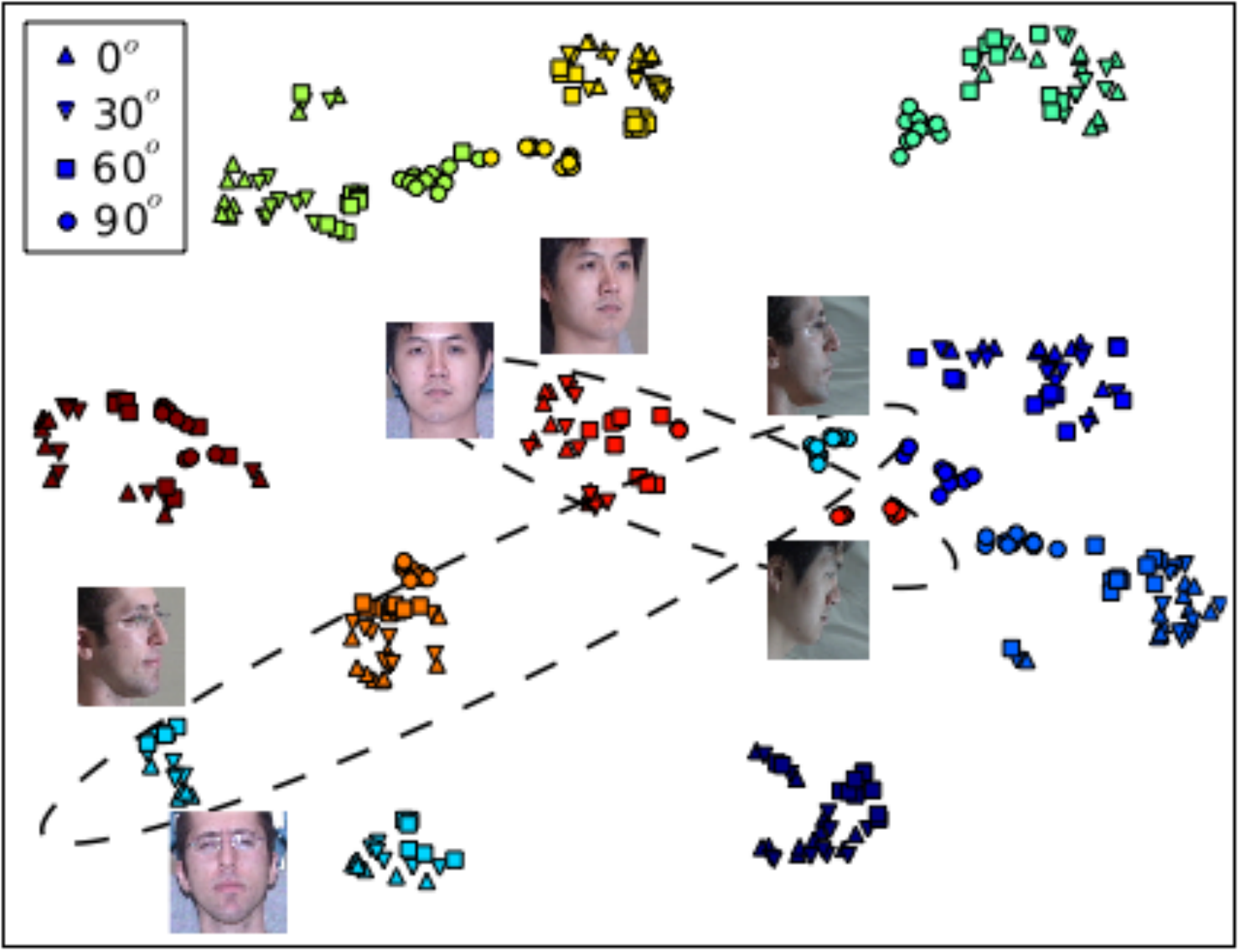}}
\subfigure[Reconstruction-based Disentangling Feature Space]{\includegraphics[width=0.453\linewidth]{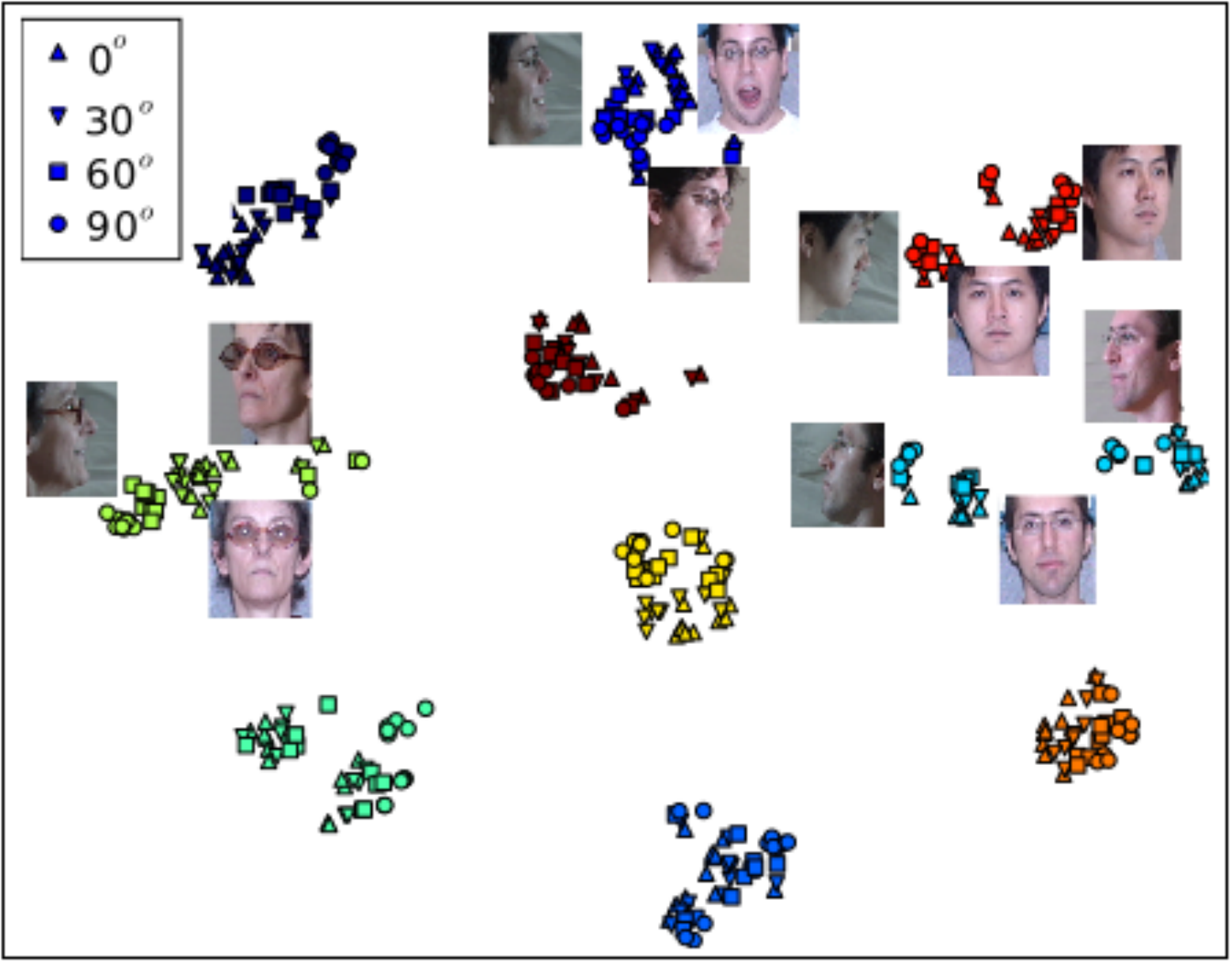}}
\caption{t-SNE visualization of VGGFace~\cite{ParkhiBMVC15} feature space (left) and the proposed reconstruction-based disentangling feature space (right), with 10 subjects from MultiPIE~\cite{multipie}. The same marker color indicates the same subject. Different marker shapes indicate different head poses. Our approach shows better results in disentangling pose factors from identity representations.} \label{fig:multipie}
\end{figure*}

Figure~\ref{fig:multipie} shows t-SNE visualization~\cite{Maaten2014} of VGGFace~\cite{ParkhiBMVC15} feature space and the proposed reconstruction-based disentangling feature space of MultiPIE~\cite{multipie}. For visualization clarity, we only visualize 10 randomly selected subjects from the test set with $0^{\circ}$, $30^{\circ}$, $60^{\circ}$, and $90^{\circ}$ yaw angles. Figure~\ref{fig:multipie} (a) shows that samples from VGGFace feature embedding have large overlap among different subjects. In contrast, Figure~\ref{fig:multipie} (b) shows that our approach can tightly cluster samples of the same subject together which leads to little overlap of different subjects, since identity features have been disentangled from pose in this case.


\subsection{Feature Embedding of 300WLP}
\begin{figure*}[h]
\centering
\subfigure[VGGFace Feature Space]{\includegraphics[width=0.42\linewidth]{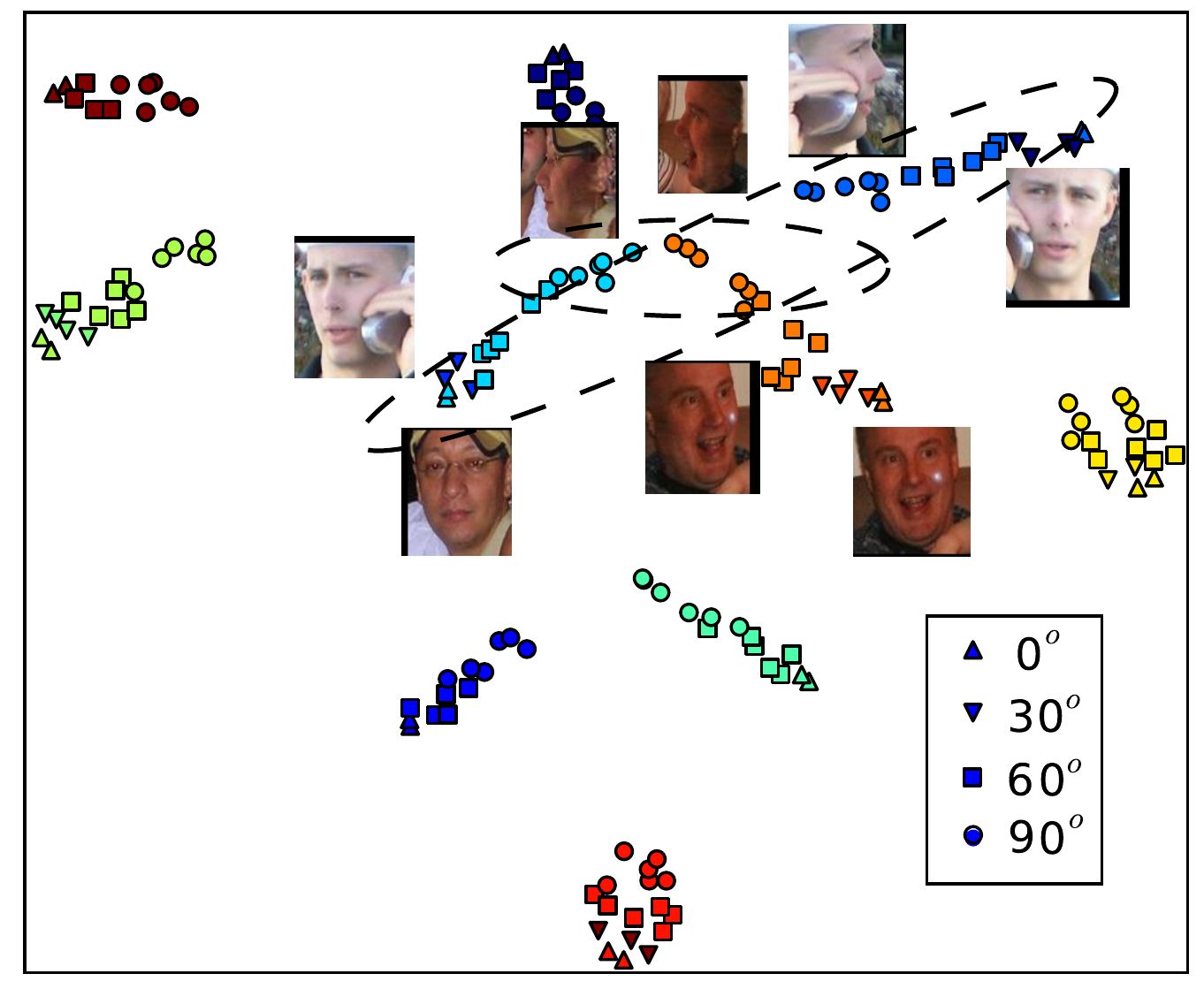}}
\subfigure[Reconstruction-based Disentangling Feature Space]{\includegraphics[width=0.505\linewidth]{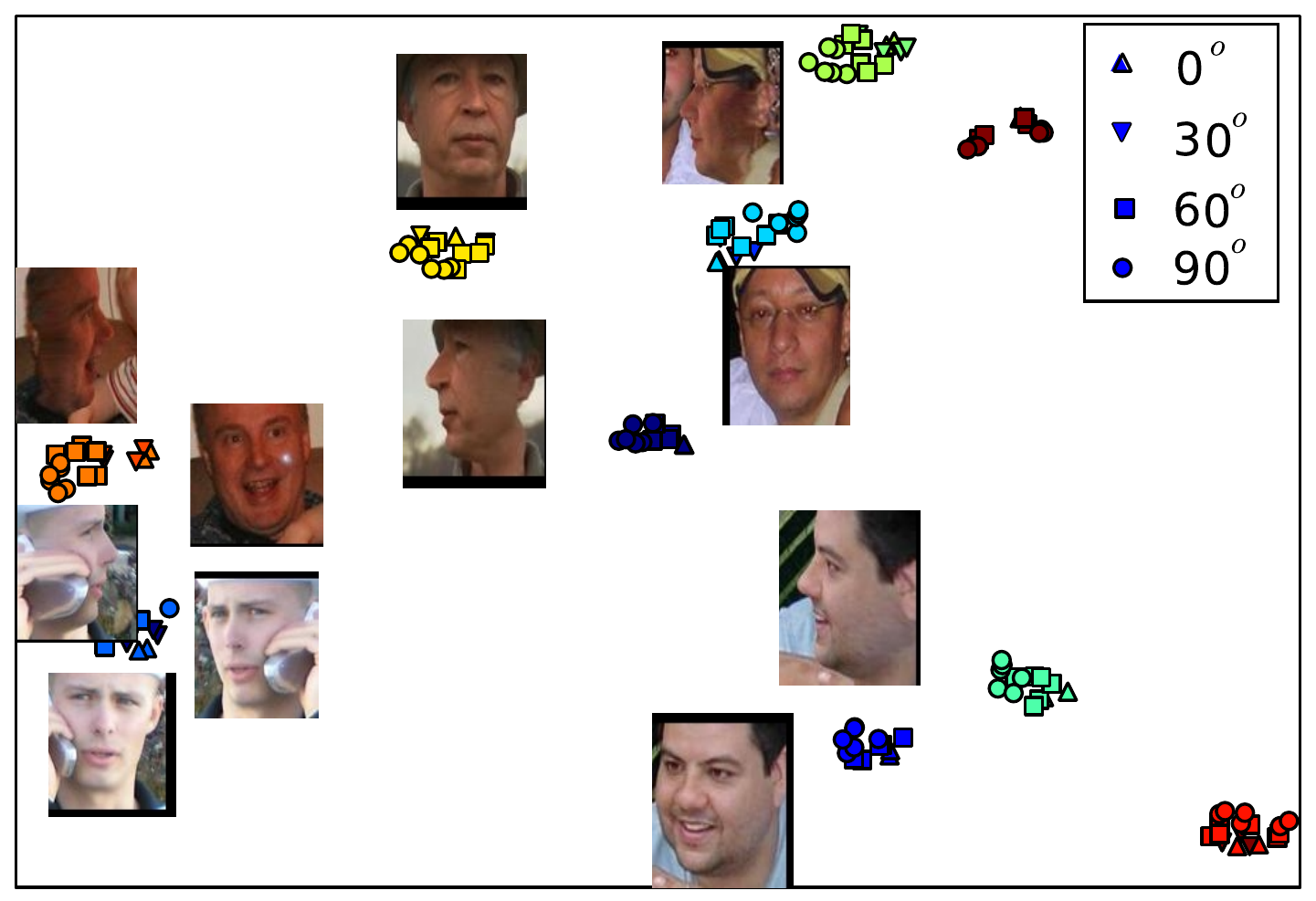}}
\caption{t-SNE visualization of VGGFace~\cite{ParkhiBMVC15} feature space (left) and the proposed reconstruction-based disentangling feature space (right), with 10 subjects from 300WLP~\cite{zhuxiangyu2016}. The same marker color indicates the same subject. Different marker shapes indicate different head poses. Our approach shows better results in disentangling pose factors from identity representations.} \label{fig:300wlp}
\end{figure*}

Figure~\ref{fig:300wlp} shows t-SNE visualization~\cite{Maaten2014} of VGGFace~\cite{ParkhiBMVC15} feature space and the proposed reconstruction-based disentangling feature space, with 10 subjects from 300WLP~\cite{zhuxiangyu2016}. Similar to the results of MultiPIE~\cite{multipie}, the VGGFace feature embedding space shows entanglement between identity and the pose, i.e., the man with the phone in $45^{\circ}$ view is overlapped with the frontal view image of other persons. In contrast, feature embeddings of our method are largely separated from one subject to another, while feature embeddings of the same subject are clustered together even there are extensive pose variations.


\subsection{Probe and Gallery Examples}
\begin{figure*}[t]
\centering
\includegraphics[width=\linewidth]{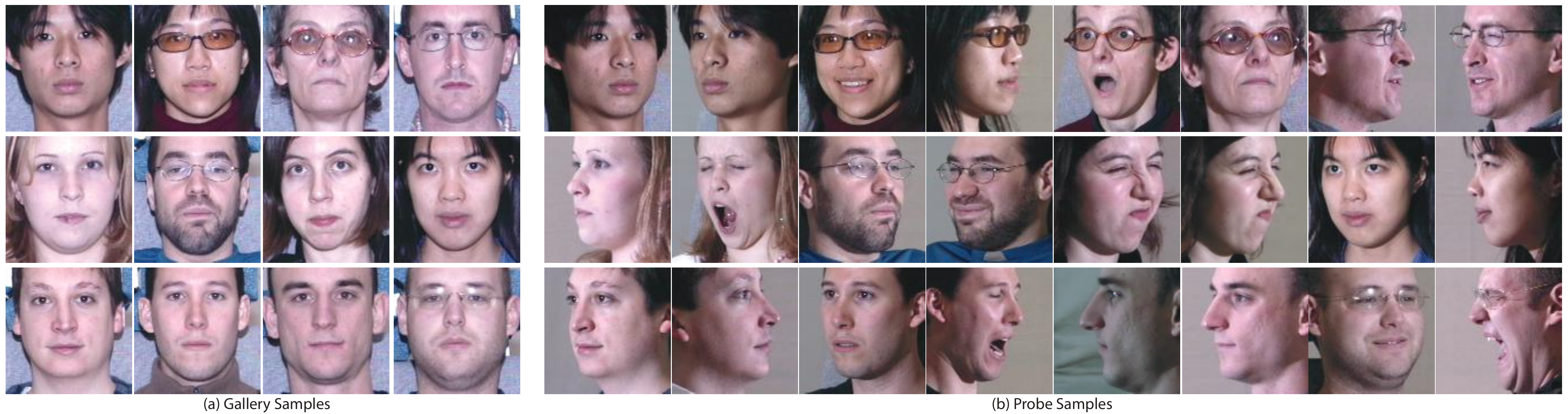}
\includegraphics[width=\linewidth]{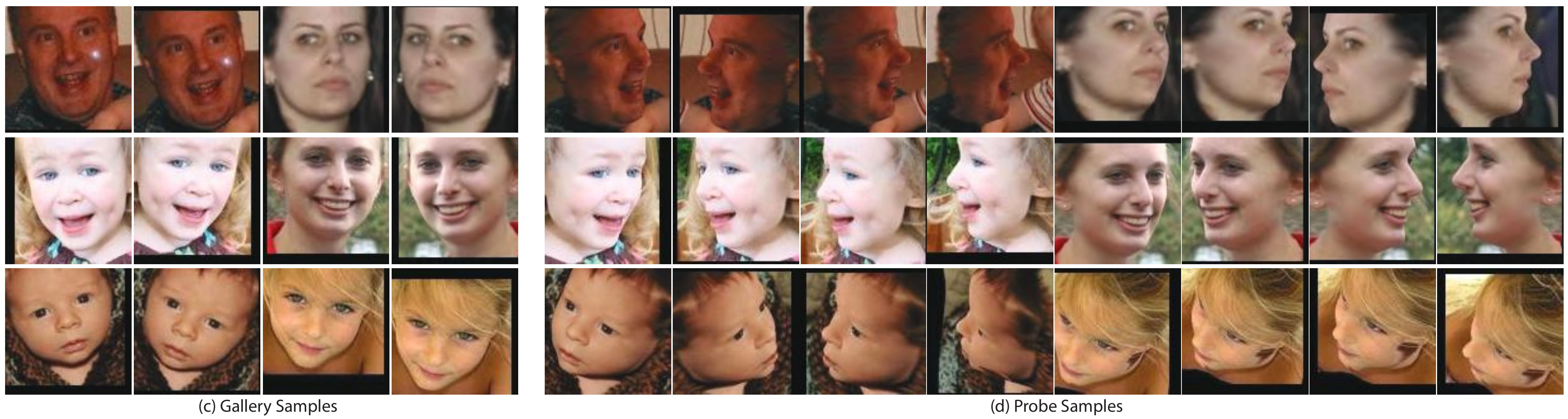}
\caption{The gallery and probe samples adopted in the testing from MultiPIE~\cite{multipie} and 300WLP~\cite{ZhuXYCVPR15}. (a) The gallery samples of MultiPIE. (b) The probe samples of MultiPIE. (c) The gallery samples of 300WLP. (d) The probe samples of 300WLP.} \label{fig:gp}
\end{figure*}

In Figure~\ref{fig:gp}, we show examples of gallery and probe images that are used in testing. Figure~\ref{fig:gp} (a) shows the gallery images in $0^{\circ}$ from MultiPIE. Each subject only has one frontal image for reference. Figure~\ref{fig:gp} (b) shows probe images of various pose and expression from MultiPIE. Each subject presents all possible poses and expressions such as neutral, happy, surprise, etc. The illumination is controlled with plain front lighting. Figure~\ref{fig:gp} (c) shows the gallery images from 300WLP, with two near-frontal images of each subject randomly selected. Figure~\ref{fig:gp} (d) shows all poses of the same subject from 300WLP. 

\subsection{Failure cases in MultiPIE and 300WLP}
\begin{figure*}[th]
\centering
\includegraphics[width=0.85\linewidth]{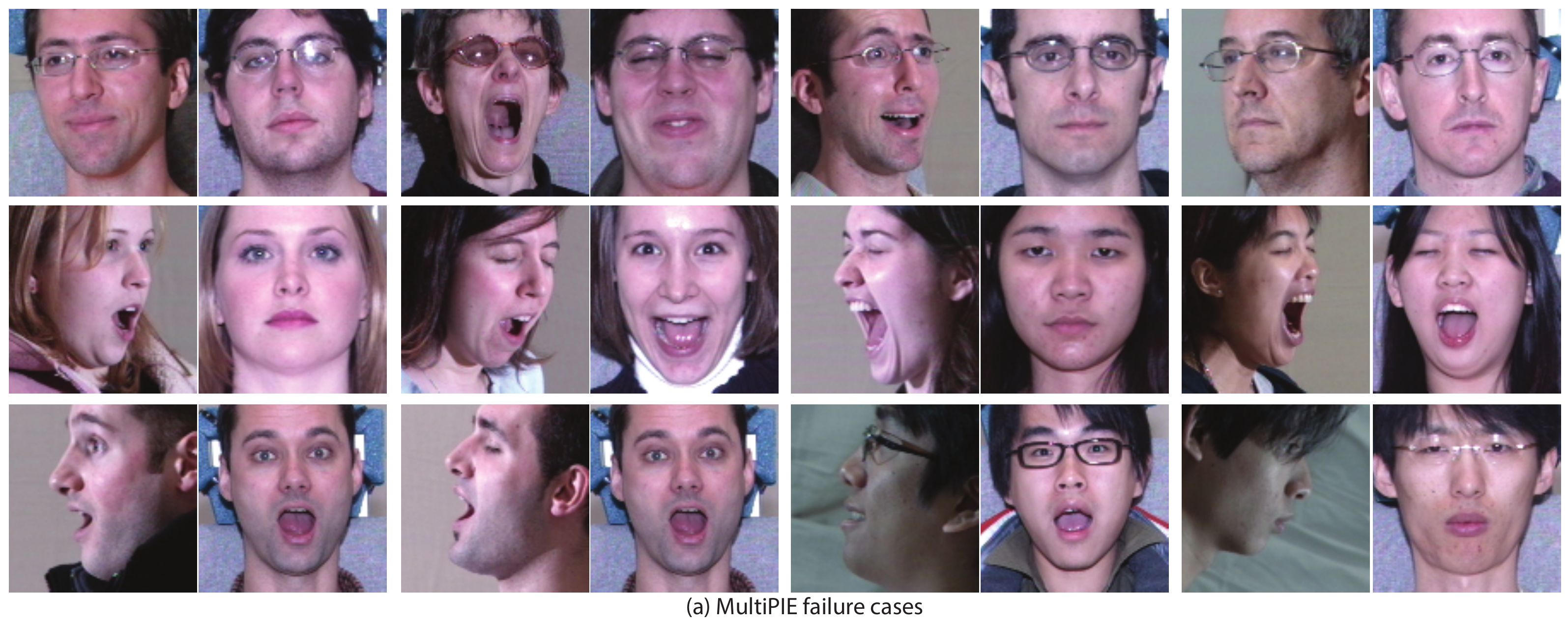}
\includegraphics[width=0.85\linewidth]{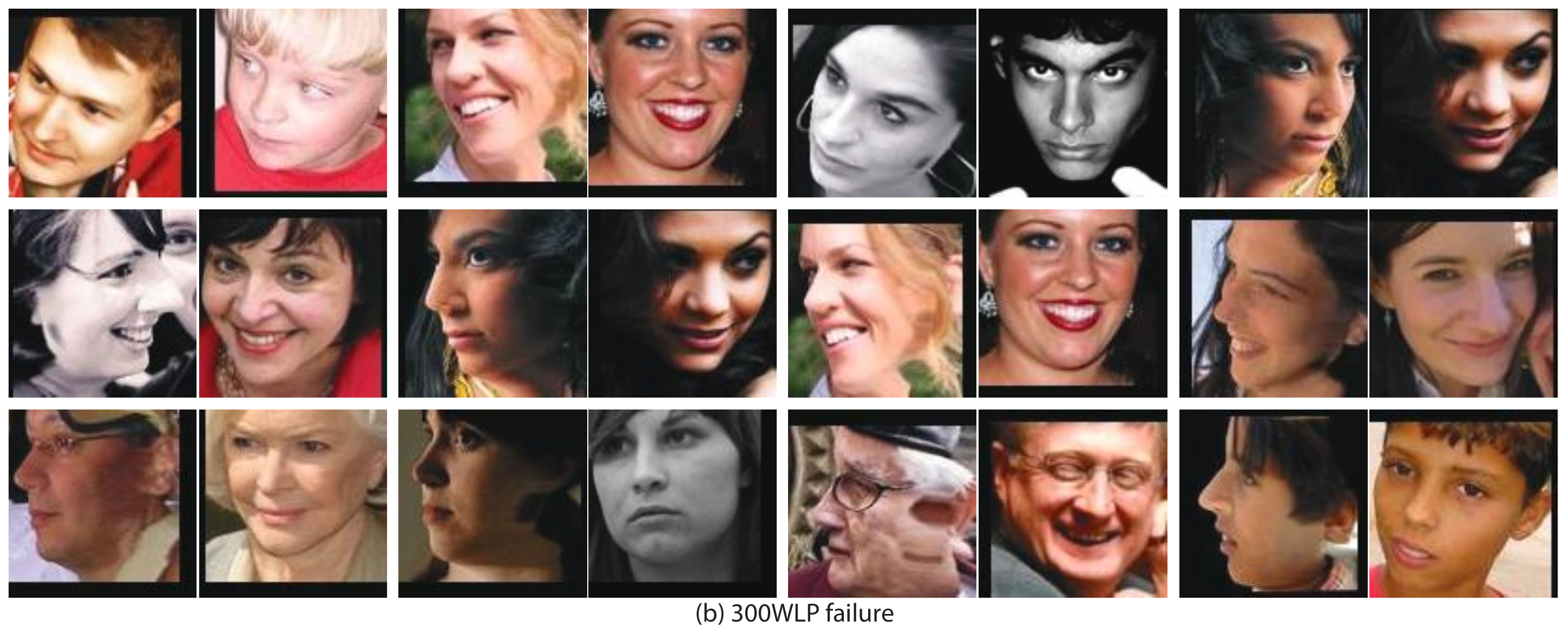}
\caption{Some failure cases in MultiPIE~\cite{multipie} and 300WLP~\cite{ZhuXYCVPR15}. Each case consists of a pair of images. The gallery image is on the left and the probe image is on the right. In both (a) and (b), the first row shows cases of $15^{\circ}$ and $30^{\circ}$, the second row shows cases of $45^{\circ}$ and $60^{\circ}$, and the third row shows cases of $75^{\circ}$ and $90^{\circ}$. (b) follows the same layout as (a). In MultiPIE, most failures result from extensive expressions. In 300WLP, most failures results from the large pose and illumination changes. Images in most failure pairs are visually similar.}
\label{fig:miss}
\end{figure*}

In Figure~\ref{fig:miss}, we show the typical failure cases generated by the proposed method on both MultiPIE and 300WLP. For MultiPIE, the most challenging cases come from exaggerated expression variations, e.g. Figure~\ref{fig:miss} (a), the second row. For 300WLP, the challenge mostly come from head pose variations and illumination variations. However, images in most failure pairs are visually similar.

\begin{table*}[t]
\begin{center}
\small
\setlength\tabcolsep{2.5pt}
\begin{tabular}{|c|c|c|c|c|c|}
\hline
\multirow{2}{*}{Method} & \multicolumn{5}{c|}{LFW}\\
\cline{2-6}
& $0-30^{\circ}$ & $30-45^{\circ}$ & $45-60^{\circ}$ & $60-90^{\circ}$ & $>30^{\circ} in avgerage$\\
\hline\hline
VGG-Face  & 0.973 (5304/5524) & 0.967 (410/424) & 0.961 (49/51) & 1.00 (1/1) & 0.964 \\
\hline
Ours & \bf{0.986} (5445/5524) & \bf{0.981} (416/424) & \bf{1.00} (51/51) & 1.00 (1/1) & \bf{0.983} \\
\hline
\end{tabular}
\end{center}
\caption{Pose-wise recognition accuracy on LFW (correct pairs num. / total pairs num. in the range).}
\label{tb:lfw}
\end{table*}

%% file: siamese_reconstruction.bbl
\begin{thebibliography}{10}\itemsep=-1pt

\bibitem{BengioLCW09}
Y.~Bengio, J.~Louradour, R.~Collobert, and J.~Weston.
\newblock Curriculum learning.
\newblock In {\em ICML}, 2009.

\bibitem{BlanzTPAMI03}
V.~Blanz and T.~Vetter.
\newblock Face recognition based on fitting a {3D} morphable model.
\newblock {\em TPAMI}, 25(9):1063--1074, 2003.

\bibitem{CaoTVCG14}
C.~Cao, Y.~Weng, S.~Zhou, Y.~Tong, and K.~Zhou.
\newblock {FaceWarehouse}: a {3D} facial expression database for visual
  computing.
\newblock {\em TVCG}, 20(3):413--425, Mar. 2014.

\bibitem{chenicip2016}
J.-C. Chen, J.~Zheng, V.~Patel, and R.~Chellappa.
\newblock Fisher vector encoded deep convolutional features for unconstrained
  face verification.
\newblock In {\em ICIP}, 2016.

\bibitem{DuongCVPR15}
C.~N. Duong, K.~Luu, K.~G. Quach, and T.~D. Bui.
\newblock Beyond principal components: Deep boltzmann machines for face
  modeling.
\newblock In {\em CVPR}, 2015.

\bibitem{Edelman_Bulthoff_1992}
S.~Edelman and H.~H. B\"ulthoff.
\newblock Orientation dependence in the recognition of familiar and novel views
  of three-dimensional objects.
\newblock {\em Vision Research}, 32(12):2385--2400, 1992.

\bibitem{multipie}
R.~Gross, I.~Matthew, J.~Cohn, T.~Kanade, and S.~Baker.
\newblock Multipie.
\newblock {\em Image and Vision Computing}, 2009.

\bibitem{Hardoon2004}
D.~Hardoon, S.~Szedmak, and J.~Shawe-Taylor.
\newblock Cannonical correlation analysis: an overview with application to
  learning methods.
\newblock {\em Neural Comput.}, 16, 2004.

\bibitem{HassnerCVPR15}
T.~Hassner, S.~Harel, E.~Paz, and R.~Enbar.
\newblock Effective face frontalization in unconstrained image.
\newblock In {\em CVPR}, 2015.

\bibitem{Hummel_Biederman_1992}
J.~E. Hummel and I.~Biederman.
\newblock Dynamic binding in a neural network for shape recognition.
\newblock {\em Psychological Review}, 99(3):480--517, 1992.

\bibitem{kan2014}
M.~Kan, S.~Shan, H.~Chang, and X.~Chen.
\newblock Stacked progressive auto-encoders (spae) for face recognition across
  poses.
\newblock In {\em CVPR}, 2014.

\bibitem{Kanmeina16}
M.~Kan, S.~Shan, and X.~Chen.
\newblock Multi-view deep network for cross-view classification.
\newblock In {\em CVPR}, 2016.

\bibitem{kan2012}
M.~Kan, S.~Shan, H.~Zhang, S.~Lao, and X.~Chen.
\newblock Multi-view discriminant analysis.
\newblock In {\em ECCV}, 2012.

\bibitem{kingma2014adam}
D.~Kingma and J.~Ba.
\newblock Adam: A method for stochastic optimization.
\newblock {\em arXiv preprint: 1412.6980}, 2014.

\bibitem{KrizhevskyNIPS12}
A.~Krizhevsky, I.~Sutskever, and G.~E. Hinton.
\newblock Imagenet classification with deep convolutional neural networks.
\newblock In {\em NIPS}, 2012.

\bibitem{NIPS2015_5851}
T.~D. Kulkarni, W.~F. Whitney, P.~Kohli, and J.~Tenenbaum.
\newblock Deep convolutional inverse graphics network.
\newblock In {\em NIPS}, 2015.

\bibitem{LiH15cvpr}
H.~Li and G.~Hua.
\newblock Hierarchical-pep model for real-world face recognition.
\newblock In {\em CVPR}, 2015.

\bibitem{shuliang16}
S.~Liang, L.~Shapiro, and I.~Kemelmacher-Shlizerman.
\newblock Head reconstruction from internet photos.
\newblock In {\em ECCV}, 2016.

\bibitem{MasiECCV16}
I.~Masi, A.~T. {a}n Tr\~{a}n, T.~Hassner, J.~T. Leksut, and G.~Medioni.
\newblock Do we really need to collect millions of faces for effective face
  recognition?
\newblock In {\em ECCV}, 2016.

\bibitem{MasiCVPR16}
I.~Masi, S.~Rawls, G.~Medioni, and P.~Natarajan.
\newblock Pose-aware face recognition in the wild.
\newblock In {\em CVPR}, 2016.

\bibitem{nelson2002}
A.~Nielson.
\newblock Multiset canonical correlations analysis and multispectral, truly
  multitemporal remote sensing data.
\newblock {\em IEEE Trans. on Image Processing}, 11(3), 2002.

\bibitem{ParkhiBMVC15}
O.~M. Parkhi, A.~Vedaldi, and A.~Zisserman.
\newblock Deep face recognition.
\newblock In {\em BMVC}, 2015.

\bibitem{PaysanAVSS09}
P.~Paysan, R.~Knothe, B.~Amberg, S.~Romdhani, and T.~Vetter.
\newblock A {3D} face model for pose and illumination invariant face
  recognition.
\newblock In {\em AVSS}, 2009.

\bibitem{peng2015circle}
X.~Peng, J.~Huang, Q.~Hu, S.~Zhang, A.~Elgammal, and D.~Metaxas.
\newblock From circle to 3-sphere: Head pose estimation by instance
  parameterization.
\newblock {\em Computer Vision and Image Understanding}, 136:92--102, 2015.

\bibitem{peng2017toward}
X.~Peng, S.~Zhang, Y.~Yu, and D.~N. Metaxas.
\newblock Toward personalized modeling: Incremental and ensemble alignment for
  sequential faces in the wild.
\newblock {\em International Journal of Computer Vision}, pages 1--14, 2017.

\bibitem{Poggio_Edelman_1990}
T.~Poggio and S.~Edelman.
\newblock A network that learns to recognize 3-dimensional objects.
\newblock {\em Nature}, 343(6255):263--266, 1990.

\bibitem{icml2014disentangling}
S.~Reed, K.~Sohn, Y.~Zhang, and H.~Lee.
\newblock Learning to disentangle factors of variation with manifold
  interaction.
\newblock In {\em ICML}, 2014.

\bibitem{rifai2012disentangling}
S.~Rifai, Y.~Bengio, A.~Courville, P.~Vincent, and M.~Mirza.
\newblock Disentangling factors of variation for facial expression recognition.
\newblock In {\em ECCV}, 2012.

\bibitem{roth15}
J.~Roth, Y.~Tong, and X.~Liu.
\newblock Unconstrained 3d face reconstruction.
\newblock In {\em CVPR}, 2015.

\bibitem{300w2013}
C.~Sagonas, G.~Tzimiropoulos, S.~Zafeiriou, and M.~Pantic.
\newblock 300 faces in-the-wild challenge: The first facial landmark
  localization challenge.
\newblock In {\em ICCVW}, 2013.

\bibitem{sanka2016}
S.~Sankaranarayanan, A.~Alavi, C.~Castillo, and R.~Chellappa.
\newblock Triplet probabilistic embedding for face verification and clustering.
\newblock In {\em arXiv preprint}, volume 1605.05396, 2016.

\bibitem{SchroffCVPR15}
F.~Schroff, D.~Kalenichenko, and J.~Philbin.
\newblock {FaceNet}: A unified embedding for face recognition and clustering.
\newblock In {\em CVPR}, 2015.

\bibitem{cfp}
S.~Sengupta, J.-C. Chen, C.~Castillo, V.~Patel, R.~Chellappa, and D.~Jacobs.
\newblock Frontal to profile face vefirication in the wild.
\newblock In {\em WACV}, 2016.

\bibitem{sharma2012}
A.~Sharma, A.~Kumar, H.~D. III, and D.~Jacobs.
\newblock Generalized multiview analysis: A discriminative latent space.
\newblock In {\em CVPR}, 2012.

\bibitem{simonyan2014}
K.~Simonyan and A.~Zisserman.
\newblock Very deep convolutional networks for large-scale image recognition.
\newblock In {\em arXiv preprint}, 2014.

\bibitem{sohn2016metric}
K.~Sohn.
\newblock Improved deep metric learning with multi-class n-pair loss objective.
\newblock In {\em NIPS}, 2016.

\bibitem{SunNIPS2014}
Y.~Sun, Y.~Chen, X.~Wang, and X.~Tang.
\newblock Deep learning face representation by joint
  identification-verification.
\newblock In {\em NIPS}, pages 1988--1996. 2014.

\bibitem{SunCVPR14}
Y.~Sun, X.~Wang, and X.~Tang.
\newblock Deep learning face representation from predicting 10,000 classes.
\newblock In {\em CVPR}, 2014.

\bibitem{TaigmanCVPR14}
Y.~Taigman, M.~Yang, M.~Ranzato, and L.~Wolf.
\newblock {DeepFace}: Closing the gap to {Human-Level} performance in face
  verification.
\newblock In {\em CVPR}, 2014.

\bibitem{TranHMM16}
A.~T. Tran, T.~Hassner, I.~Masi, and G.~G. Medioni.
\newblock Regressing robust and discriminative 3d morphable models with a very
  deep neural network.
\newblock {\em CoRR}, abs/1612.04904, 2016.

\bibitem{drgan}
L.~Tran, X.~Yin, and X.~Liu.
\newblock Disentangled representation learning gan for pose-invariant face
  recognition.
\newblock In {\em CVPR}, 2017.

\bibitem{wang2017leveraging}
X.~Wang, G.~Guo, M.~Merler, N.~C. Codella, M.~Rohith, J.~R. Smith, and
  C.~Kambhamettu.
\newblock Leveraging multiple cues for recognizing family photos.
\newblock {\em Image and Vision Computing}, 58:61--75, 2017.

\bibitem{wenyandong2016}
Y.~Wen, K.~Zhang, Z.~Li, and Y.~Qiao.
\newblock A discriminative feature learning approach for deep face recognition.
\newblock In {\em ECCV}, 2016.

\bibitem{YiCoRR14}
D.~Yi, Z.~Lei, S.~Liao, and S.~Z. Li.
\newblock Learning face representation from scratch.
\newblock In {\em CoRR}, 2014.

\bibitem{xiang_ffgan}
X.~Yin, X.~Yu, K.~Sohn, X.~Liu, and M.~Chandraker.
\newblock Towards large-pose face frontalization in the wild.
\newblock In {\em ICCV}, 2017.

\bibitem{xiang_pami_cdm_2015}
X.~Yu, J.~Huang, S.~Zhang, and D.~N. Metaxas.
\newblock Face landmark fitting via optimized part mixtures and cascaded
  deformable model.
\newblock {\em IEEE Transactions on Pattern Analysis and Machine Intelligence},
  38(11):2212 -- 2226, 2015.

\bibitem{xiang_cor_2014}
X.~Yu, Z.~Lin, J.~Brandt, and D.~N. Metaxas.
\newblock Consensus of regression for occlusion-robust facial feature
  localization.
\newblock In {\em ECCV}, 2014.

\bibitem{xiangeccv16}
X.~Yu, F.~Zhou, and M.~Chandraker.
\newblock Deep deformation network for object landmark localization.
\newblock In {\em ECCV}, 2016.

\bibitem{zhuxiangyu2016}
X.~Zhu, Z.~Lei, X.~Liu, H.~Shi, and S.~Li.
\newblock Face alignment across large poses: A 3d solution.
\newblock In {\em CVPR}, 2016.

\bibitem{ZhuXYCVPR15}
X.~Zhu, Z.~Lei, J.~Yan, D.~Yi, and S.~Z. Li.
\newblock High-fidelity pose and expression normalization for face recognition
  in the wild.
\newblock In {\em CVPR}, 2015.

\bibitem{zhuzhenyao2013}
Z.~Zhu, P.~Luo, X.~Wang, and X.~Tang.
\newblock Deep learning identity-preserving face space.
\newblock In {\em ICCV}, 2013.

\bibitem{Zhu14nips}
Z.~Zhu, P.~Luo, X.~Wang, and X.~Tang.
\newblock Multi-view perceptron: a deep model for learning face identity and
  view representations.
\newblock In {\em NIPS}, 2014.

\end{thebibliography}
